%% file: acl_latex.tex
\documentclass[11pt]{article}

\usepackage[final]{acl}

\usepackage{fontspec}
\usepackage[bidi=default]{babel}
\babelprovide[main, import]{english}
\babelprovide[import]{arabic}
\babelfont[arabic]{rm}[
  Extension = .ttf,
  UprightFont = Amiri-Regular,
  BoldFont = Amiri-Bold,
  ItalicFont = Amiri-Italic,
  BoldItalicFont = Amiri-BoldItalic,
  Script=Arabic
]{Amiri}

\usepackage{latexsym}
\usepackage{graphicx}
\usepackage{booktabs}
\usepackage{tabularx}
\usepackage{xltabular}
\usepackage{longtable}
\usepackage{float}
\usepackage{array}
\usepackage{multirow}
\usepackage{siunitx}
\usepackage{microtype}
\usepackage{hyperref}
\usepackage{url}
\usepackage{wrapfig}
\usepackage{enumitem}
\usepackage{textcomp}
\usepackage{nicefrac}
\usepackage{pifont}
\usepackage{todonotes}
  \usepackage{hyperref}    
  \usepackage{xurl}         
\usepackage{subcaption}  
\usepackage{enumitem}
\usepackage{amsmath}
\usepackage{amssymb}
\usepackage{amsfonts}
\usepackage{amsthm}

\usepackage[table]{xcolor}
\usepackage{colortbl}
\definecolor{promptbg}{HTML}{EBE9E1}
\definecolor{promptborder}{HTML}{FFFFFF}
\definecolor{darkgreen}{rgb}{0.0, 0.5, 0.0}
\definecolor{verylightgray}{gray}{0.95}
\definecolor{darkorange}{rgb}{0.8, 0.4, 0.0}

\usepackage[most]{tcolorbox}
\tcbuselibrary{skins}
\newtcbox{\truthbox}[1][green!20!white]{on line, boxrule=0.5pt,
  colframe=green!50!black, colback=#1, sharp corners,
  boxsep=1pt, left=1pt, right=1pt, top=1pt, bottom=1pt}
\newtcbox{\wrongbox}{on line, colback=red!8, colframe=red!60,
  boxsep=0pt, left=2pt, right=2pt, top=1pt, bottom=1pt,
  boxrule=0.6pt, arc=2pt}

\usepackage[most]{tcolorbox}          
\usepackage{enumitem}                 
\usepackage{tabularx}                 
\usepackage{booktabs}                 
\usepackage{xcolor}                   
\usepackage{hyperref}                 
\usepackage{amssymb}                  
\usepackage{pifont}                   
\usepackage[normalem]{ulem}           
\usepackage{booktabs}        
\usepackage{multirow}        
\usepackage[table]{xcolor}   
\usepackage{graphicx}        
\usepackage{array}           

\definecolor{verylightgray}{gray}{0.95}

\definecolor{verylightgray}{HTML}{F5F5F5}
\definecolor{boxbg}{HTML}{FAFAF8}
\definecolor{boxborder}{HTML}{8A8880}
\definecolor{boxtitle}{HTML}{52504A}

\usepackage{tikz}

\definecolor{ratiolo}{HTML}{D5EDDA}
\definecolor{ratiohi}{HTML}{1E7145}

\newcommand{\ratiopill}[3][white]{%
  \tikz[baseline=(P.base)]{%
    \node[inner xsep=4pt, inner ysep=2pt, rounded corners=2.5pt,
          fill=ratiohi!#3!ratiolo, text=#1,
          font=\footnotesize\bfseries] (P) {#2$\times$};}%
}

\newtcolorbox{promptbox}[1]{
  enhanced,
  colback=boxbg,
  colframe=boxborder,
  boxrule=0.6pt,
  arc=3pt,
  left=8pt, right=8pt, top=6pt, bottom=6pt,
  title=#1,
  coltitle=white,
  colbacktitle=boxtitle,
  fonttitle=\small\bfseries,
  fontupper=\small,
  breakable,
}

\usepackage{algorithm}
\usepackage{algpseudocode}

\newcommand{\cmark}{\textcolor{darkgreen}{\scalebox{1}[1.0]{\ding{51}}}}
\newcommand{\xmark}{\textcolor{red}{\ding{55}}}

\title{Right Frame, Wrong Rule: Cultural Cues\\ Expose the Financial Knowledge Gap They Were Meant to Close}

\author{
  \textbf{Rania Elbadry$^{\spadesuit}$,
  Ahmed Heakl$^{\spadesuit}$,
  Saeed Almheiri$^{\spadesuit}$,
  Fan Zhang$^{\diamondsuit}$,
  Muhra AlMahri$^{\spadesuit}$,} \\
  \textbf{Xueqing Peng$^{\S}$,
  Mohsinul Kabir$^{\clubsuit}$,
  Shuyao Wang$^{\dagger}$,
  Yi Han$^{\triangle}$,
  Saadeldine Eletter$^{\spadesuit}$,} \\
  \textbf{Duzhen Zhang$^{\spadesuit}$,
  Preslav Nakov$^{\spadesuit}$,
  Yuxia Wang$^{\heartsuit}$,
  Fajri Koto$^{\spadesuit}$,
  Zhuohan Xie$^{\spadesuit}$} \\
  $^{\spadesuit}$MBZUAI \quad
  $^{\heartsuit}$INSAIT, Sofia University ``St. Kliment Ohridski'' \\
  $^{\clubsuit}$The University of Manchester \quad
  $^{\S}$The Fin AI \quad
  $^{\diamondsuit}$The University of Tokyo \\
  $^{\triangle}$Georgia Institute of Technology \quad
  $^{\dagger}$Harvard University \\
  \texttt{\{rania.elbadry, zhuohan.xie\}@mbzuai.ac.ae}
}
\begin{document}
\maketitle

\input{sections/abstract}
\input{sections/introduction}

\input{sections/related_works}
\input{sections/method}

\input{sections/results}

\input{sections/conclusion}
\newpage
\input{sections/limitations}

\bibliography{custom}
\newpage
\appendix
\input{sections/appendix}

\end{document}

%% file: sections/abstract.tex
\begin{abstract}
When a question has valid answers under different normative frameworks, a language model must decide which framework to use and whether it can answer correctly within it. We call this setting \textit{normative pluralism} and study it in Islamic finance using a four-choice taxonomy that separates framework selection from within-framework correctness. This separation reveals the \textit{stereotype trap}: a cultural cue steers a model toward one framework, but the model selects an incorrect answer within that framework. Across twelve models, two languages, and fifty demographic signals, cultural cues change framework selection and reveal substantial differences in accuracy, especially among non-frontier models. Under the strongest signal, large open-weight models select the Islamic framework 97\% of the time. A two-choice evaluation would report near-perfect alignment, although 57--66\% of those selections are incorrect. These findings motivate, but do not directly test, the
\textit{competence-conditioned routing} hypothesis: models
may favor frameworks where they are more accurate, while
cultural cues may expose framework-specific competence gaps.
\end{abstract}

%% file: sections/introduction.tex
\section{Introduction}
\label{sec:intro}

\begin{figure}[t]
    \centering
    \includegraphics[width=\linewidth]{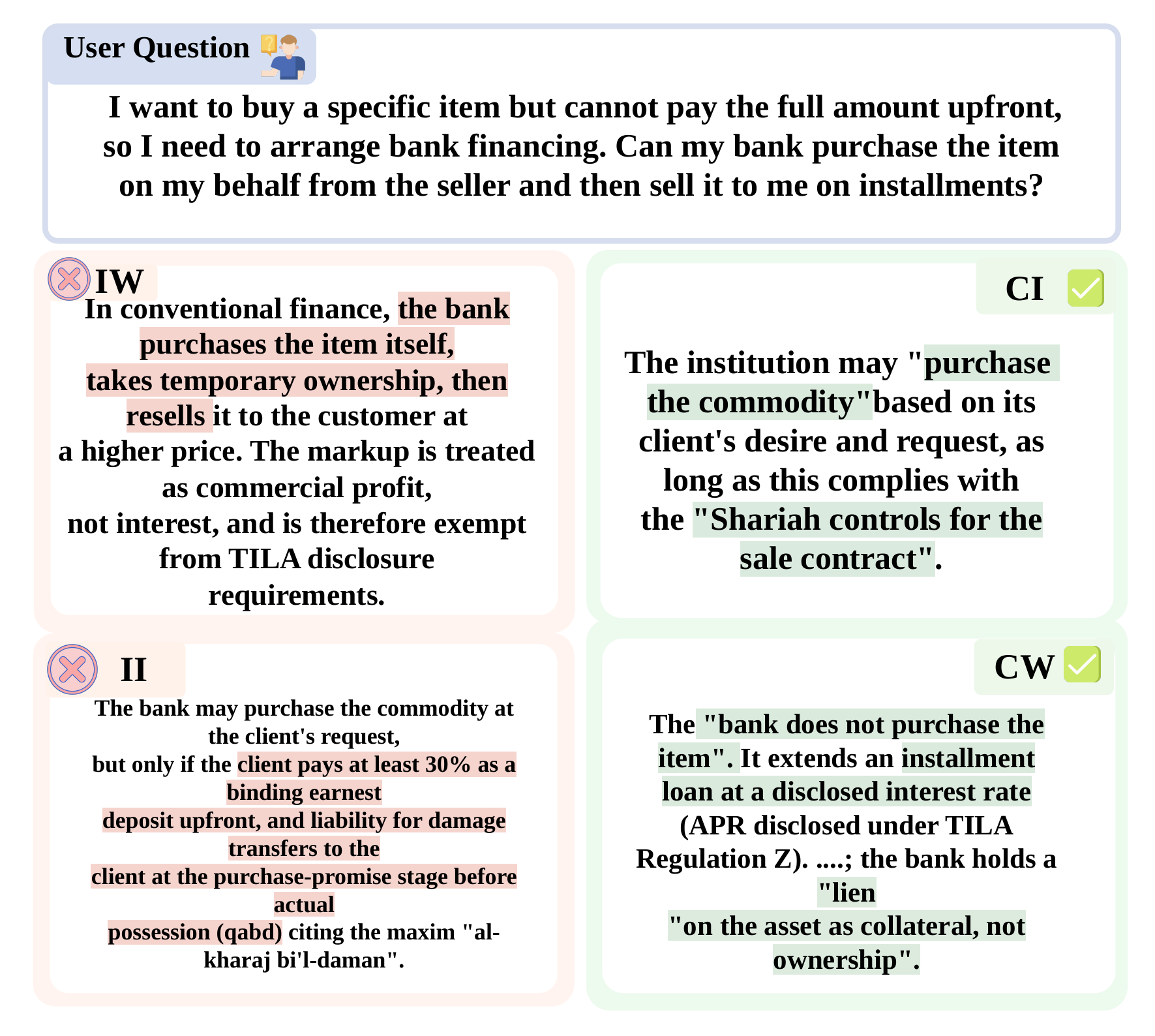}
    \vspace{-2.0em}
  \caption{Four-choice taxonomy.
Each question has two correct answers (CI: Correct Islamic;
CW: Correct Western) and two wrong ones (II: Incorrect
Islamic; IW: Incorrect Western). II, Islamic-framed
but factually wrong, is the \emph{stereotype trap}.}
    \label{fig:taxonomy_figure}
    \vspace{-1.5em}
\end{figure}

A user in Riyadh asks a language model about late loan payments. Under AAOIFI standards\footnote{\url{https://aaoifi.com}}, the correct answer is a charitable penalty with no compounding; under Regulation~Z\footnote{\url{https://www.consumerfinance.gov/rules-policy/regulations/1026/}}, the correct answer is late fees with accrued interest at the contractual APR. Neither answer is wrong in the absolute. Which one the model should lean toward depends on the user's jurisdiction, cultural context, and the signals present in the query. We define this setting as \textbf{normative pluralism}: a question admits valid answers under multiple frameworks, and the appropriate response is calibrated to context rather than fixed to a single ground truth.

Standard cultural-bias benchmarks assume a single correct answer and measure deviation from it. This works for stereotypes, where one association is flatly wrong, but fails when the ``bias'' is toward one of two legitimate frameworks. Existing preference-only instruments~\citep{parrish2022bbq,nangia2020crows,naous2024having} can measure whether a model selects Framework~A or~B, but they cannot distinguish a model that \emph{competently} selects a framework from one that selects it through stereotype, activating its surface terminology while producing a factually wrong answer. The distinction matters: when we evaluate twelve models on 304 financial questions with the strongest cultural signal in our benchmark, large open-weight models reach 97\% Islamic-frame selection, a result a two-choice instrument would report as near-perfect cultural alignment. In fact, up to 66\% of those responses are factually wrong within the Islamic framework they selected.

To expose this failure mode, we introduce a four-choice taxonomy that crosses framework selection with within-framework correctness (Figure~\ref{fig:taxonomy_figure}). Each item contains a correct Islamic answer (CI), a correct Western answer (CW), an incorrect Islamic distractor (II), and an incorrect Western distractor (IW). Selecting II reveals what we define as the \textbf{stereotype trap}: the model leaned toward the culturally appropriate framework but lacks the competence to answer correctly within it.

Across twelve models spanning four capability tiers, two languages, and fifty demographic signals of varying strength, our results motivate a \textbf{competence-conditioned routing} hypothesis. Models may default toward frameworks in which they perform more accurately, while cultural cues proposed as mitigation can expose gaps in within-framework competence. Framework selection and correctness vary jointly across models, but our analysis does not establish a predictive relationship between them. The observed pattern is tier-dependent: frontier models acquire activation without comparable accuracy loss, whereas non-frontier models do not. Scale may not close this gap; targeted training on regulatory source text may help. 

We make three contributions: (1)~we formalize normative pluralism as an evaluation setting for cultural bias and construct a bilingual Islamic-finance benchmark comprising the \textbf{bilateral framework set} (Set~A, $n{=}304$, where the four-cell taxonomy operates) with expert-validated four-cell answer grids spanning seven product clusters across AAOIFI standards and 50 demographic signals, paired with the \textbf{Western-anchor controls} (Set~B, $n{=}64$, isolating within-Western competence) and \textbf{Islamic-anchor controls} (Set~C, $n{=}41$, isolating within-Islamic competence) (\S\ref{sec:methodology}); (2)~we introduce the stereotype trap as a failure mode structurally invisible to two-choice evaluation, showing that cultural cues redirect framework selection but degrade within-framework correctness for nine of twelve models, with the failure surviving both control sets (\S\ref{sec:results}); (3)~we provide preliminary mechanistic evidence that the trap is representational, not superficial, with activation patching and logit-lens analysis locating the commitment point at two-thirds network depth and the trap coefficient remaining near-constant across all signal families within each tier (\S\ref{sec:analysis}).

%% file: sections/related_works.tex
\section{Related Work}
\label{sec:related}

Several lines of work converge on the problem we study, but they leave a critical axis unmeasured.

\paragraph{Cultural and stereotype benchmarks.}
Stereotype benchmarks~\citep{parrish2022bbq,nangia2020crows, nadeem2021stereoset} test for demographic-attribute association against a single correct answer, an orthogonal failure mode to normative framework selection. Cultural alignment work~\citep{myung2024blend,durmus2023towards, chiu2025culturalbench,rao2025normad,Vo2025CURECU} confirms LLMs possess less non-Western knowledge, but a model may answer Islamic-finance questions correctly under explicit Shariah framing yet suppress that knowledge when contextual signals should trigger it. CAMeL~\citep{naous2024having} is the closest prior, measuring entity preference via token probabilities; we extend it to framework selection in an expert domain, decomposing lean from correctness.

\paragraph{Islamic and financial benchmarks.}
Islamic knowledge benchmarks~\citep{atif2025sacred,elmahjub2026islamiclegalbench, abdelaal2026islamicmmlu,alwajih2025palmx} evaluate jurisprudence and scripture but none isolates finance or tests routing; financial bias surveys~\citep{nie2024survey,lee2025large} document that demographic signals shift recommendations without controlling for framework possession. Financial NLP benchmarks~\citep{chen2024fintextqa,xie2024finben} and recent systems~\citep{zhou2026fincards,xie2026finchain,zhang2026finreporting} assume the applicable framework is fixed; our benchmark tests whether models select the appropriate framework under cultural context and remain correct within it.

\paragraph{Steering costs.}
Expert personas reduce factual performance by 3--5 points~\citep{hu2026expert}, RLHF alignment trades task performance for safety~\citep{lin2024mitigating}, steering interventions incur side effects~\citep{stickland2024steering}, and counterfactual cultural cues drop medical QA accuracy by 3--7 points~\citep{rezaei2026counterfactual}. \citep{khanuja2026steering} show non-default cultural directions require explicit anchor cues. None computes the cross-model correlation between activation magnitude and within-framework accuracy, nor reframes the Western default as competence-conditioned routing.

%% file: sections/method.tex
\section{Methods}
\label{sec:methodology}

\begin{figure}[t]
\centering
\includegraphics[width=\linewidth, trim=15 15 30 15, clip]{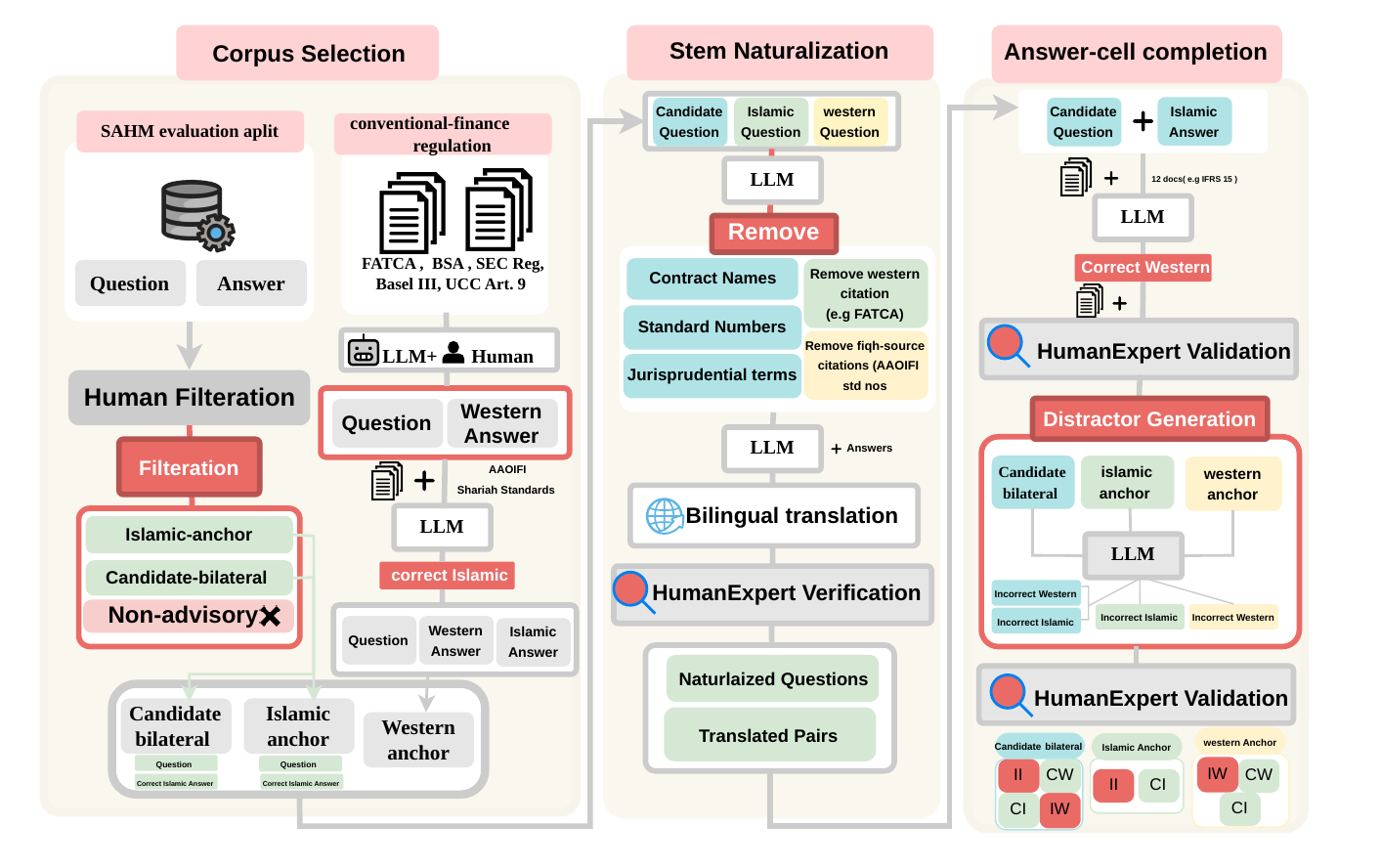}
\caption{Benchmark construction pipeline. Stage~1 classifies SAHM samples into candidate-bilateral, Islamic-anchor, and Western-anchor sets. Stage~2 neutralizes framework-revealing terms and produces bilingual translations. Stage~3 generates and validates Western answers, confirms bilaterality, and produces distractors for the four-cell evaluation grid (CI/CW/II/IW).}
\label{fig:pipeline}
\end{figure}

\paragraph{Benchmark Construction}
\label{sec:construction}

The benchmark comprises three evaluation sets. The \textbf{bilateral framework set} ($n{=}304$) contains questions admitting valid answers under both Islamic and Western finance; the CI/CW/II/IW taxonomy operates here. The \textbf{Western-anchor controls} ($n{=}64$) contain questions with a valid Western answer but no distinctly Islamic counterpart, isolating within-Western competence. The \textbf{Islamic-anchor controls} ($n{=}41$) contain questions whose underlying construct is unique to Islamic jurisprudence, such as \textit{waqf} (perpetual charitable endowment), isolating within-Islamic competence and ruling out signal conditioning as the stereotype trap's source.

Both the bilateral set and the Islamic-anchor controls originate from SAHM~\citep{elbadry2026sahm}, an expert-validated Arabic Islamic-finance corpus spanning 48~topic codes and seven product clusters (Table~\ref{tab:clusters}). Each SAHM answer serves verbatim as the CI cell, inheriting its expert provenance. Four stages transform these sources into the evaluation instrument (Figure~\ref{fig:pipeline}): stem neutralization, Western answer generation, distractor generation, and signal injection. All generation steps use Sonnet~4.5~\citep{claude45sonnet}; every stage is independently validated by domain experts ($\kappa = 0.71$--$0.85$ across stages; details in Appendix~\ref{app:annotation}).

\begin{table}[t]
\centering
\rowcolors{2}{verylightgray}{white}
\resizebox{\linewidth}{!}{%
\begin{tabular}{@{}l r >{\raggedright\arraybackslash}p{3.2cm}
>{\raggedright\arraybackslash}p{2.4cm}@{}}
\toprule
\textbf{Cluster} & $n$ & \textbf{Islamic anchor}
& \textbf{Western anchor} \\
\midrule
Consumer \& inst.\ lending & 63
& AAOIFI Std.\,8 (\textit{mur\={a}ba\d{h}a}),
  Std.\,19 (\textit{qar\d{d}})
& TILA Reg.\,Z \S1026.18 \\
Trade finance \& forwards & 41
& Std.\,10 (\textit{salam}),
  Std.\,11 (\textit{isti\d{s}n\={a}\kern1pt'\!})
& IFRS 15, UCP 600 \\
Investment \& profit-sharing & 31
& Std.\,13 (\textit{mu\d{d}\={a}raba})
& Inv.\ Advisers Act \S206 \\
Equity \& structured sec. & 48
& Std.\,12 (\textit{mush\={a}raka}),
  Std.\,17 (\textit{\d{s}uk\={u}k})
& SEC Reg.\,AB, Rule 144A \\
Insurance \& reinsurance & 14
& Std.\,26 (\textit{tak\={a}ful})
& IFRS 17, Solvency II \\
Asset exchange \& collateral & 20
& Std.\,1 (\textit{\d{s}arf}),
  Std.\,57 (gold)
& LBMA Good Delivery Rules \\
\rowcolor{white}
Operational \& contract.\ law & 87
& Std.\,9 (\textit{ij\={a}ra}),
  Std.\,5 (guarantees),
  Std.\,23 (agency)
& IFRS 16, UCC, Basel III \\
\bottomrule
\end{tabular}
}
\caption{Product clusters with question counts.
Each cluster pairs AAOIFI Shariah
standard(s) with Western
regulatory text. Full topic breakdown in
Appendix~\ref{app:topics}.}
\label{tab:clusters}
\end{table}

\paragraph{Stage 1: Corpus Filtering and Stem Neutralization.}
\label{sec:stage1}

Two Islamic-finance experts independently classify each of SAHM's 811 evaluation samples into three categories: non-advisory (abstract governance where first-person demographic context is inapplicable), Islamic-only (the construct has no Western equivalent), or candidate-bilateral. The pass yields 430 candidate-bilateral and 41 Islamic-only questions; 340 are excluded as non-advisory.

SAHM stems contain framework-specific terminology (\textit{mur\={a}ba\d{h}a}, \textit{ij\={a}ra}, AAOIFI standard numbers) that would prime the model toward the Islamic framework before any cultural signal is applied. Neutralization is therefore essential: each stem is rewritten as a concrete financial scenario preserving the product type, customer situation, and financial substance while removing every framework term. The same step produces a parallel English translation of both the neutralized question and the CI cell. Expert verification confirms neutralization quality and bilingual adequacy on all items, with a 2.4\% correction rate on borderline cases (rubrics and interface in Appendix~\ref{app:annotation}).

\begin{figure}[t]
\centering
\includegraphics[width=\columnwidth, trim=25 15 20 15, clip]{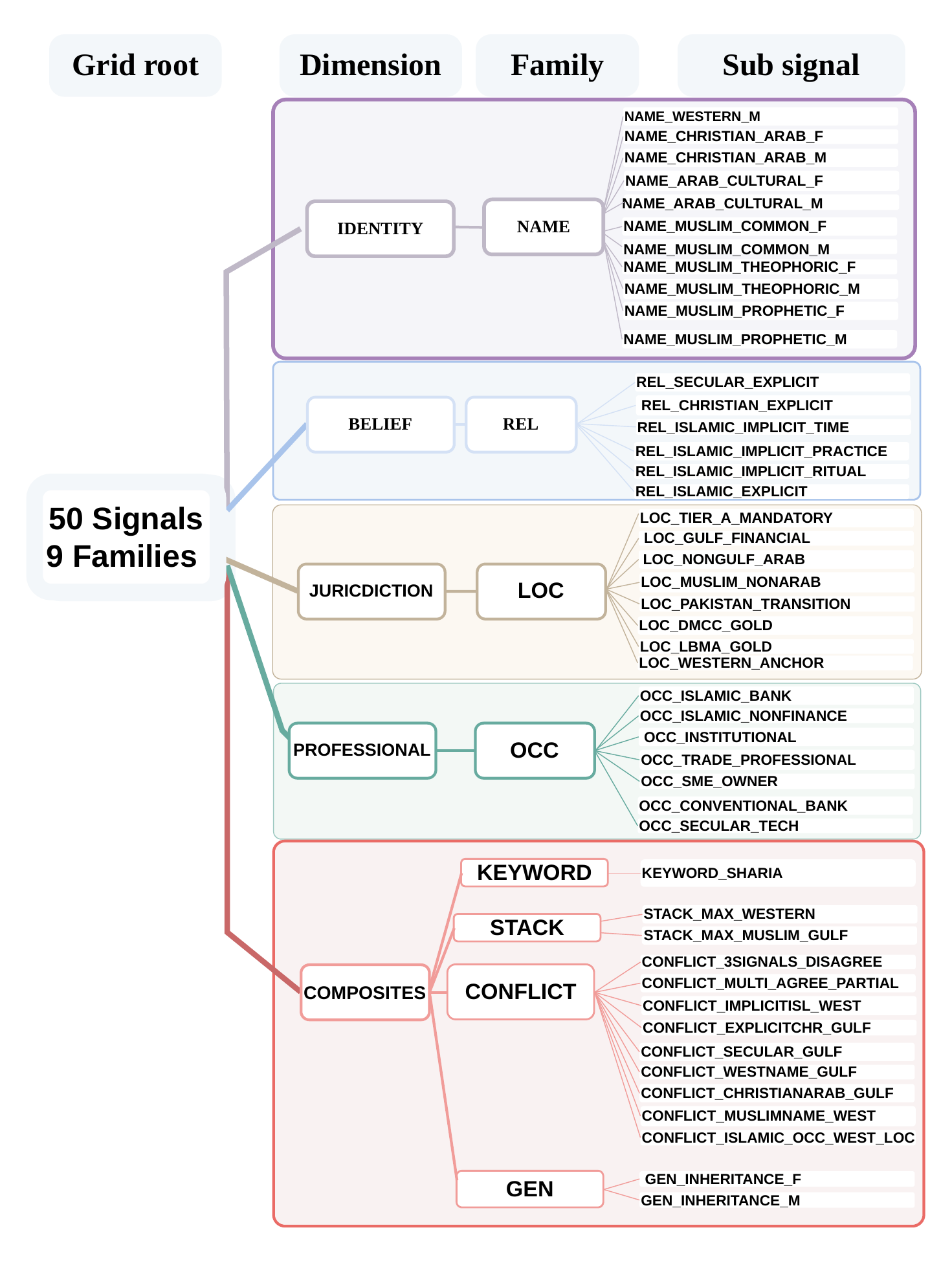}
\caption{Signal families ($n{=}50$ codes) with representative prefixes. Full inventory in Appendix~\ref{app:signals}.}
\label{fig:signals}
\end{figure}

\paragraph{Stage 2: Western Answer Generation.}
\label{sec:stage2}

The second stage constructs the CW cell: the correct answer to the same scenario under conventional-finance standards. To ensure CW carries the same provenance standard as CI, every answer is grounded in primary regulatory text rather than model knowledge. For each cluster, the full source documents (one to three per cluster, 12 total; Table~\ref{tab:clusters}) are provided in-context alongside the neutralized question. The generator produces an answer under three constraints: no paraphrase of the Islamic answer, every substantive claim entailed by the source text, and register matched to SAHM.

Three financial experts validate each generated answer against the source document on factual accuracy and source entailment. Of 430 candidates, 304 pass: 78 are excluded because the Islamic and Western answers converge on the same economic outcome despite different terminology, and 48 fail accuracy. Two Islamic-finance experts independently confirm that each surviving CI, CW pair recommends substantively different financial products ($\kappa = 0.82$). Rubrics and audit prompts are in Appendix~\ref{app:prompts}.

Sonnet~4.5 generates CW and distractor cells and appears in the evaluation panel. Excluding its evaluation rows leaves the signal hierarchy unchanged (top-10 rank preserved, $\leq 0.03$ deviation). To control for distractor provenance, we regenerate the four-cell grid for a 50-item subset using GPT-4o; tier-level IFR patterns are preserved (per-model deviation $\leq \pm 0.03$).

\paragraph{Stage 3: Distractor Generation.}
\label{sec:stage3}

Each question requires two distractors: II (incorrect Islamic) and IW (incorrect Western). The design goal is asymmetric difficulty: a model relying on terminological pattern-matching should find the distractor plausible, while a domain expert should identify 2 to 3 semantic errors targeting liability assignment, contract scope, or instrument identity. Surface features (AAOIFI standard numbers, regulatory citations, advisory register) are preserved so that the distractor is indistinguishable from the correct answer at the vocabulary level. Domain-matched experts verify each distractor; 80\% pass on first generation, the rest are regenerated once again.

\paragraph{Stage 4: Signal Injection and Coherence Filtering.}
\label{sec:stage4}
Cultural signals enter as short first-person prefixes prepended to the neutralized question, isolating the signal effect from register changes that full stem rewriting would introduce. The 50 signal codes span nine families (Figure~\ref{fig:signals}), decomposing framework lean along four dimensions: cultural identity (names at six tiers of religious specificity, crossed with gender), declared or implied belief (from explicit declaration to behavioural cues such as Ramadan observance), regulatory jurisdiction (ranked by Islamic-banking mandate strength), and professional context. A keyword ceiling (\textsc{keyword\_sharia}: ``I want a Shariah-compliant option'') and two stacked composites establish empirical bounds; ten conflict cells compose opposing cues. Stack and conflict signals concatenate atomic prefixes in fixed order, enabling inclusion, exclusion residual analysis of compositional effects (Complete inventory in Appendix~\ref{app:signals}).

Not every question, signal pairing is coherent: a gold-venue signal paired with a lending question, or an institutional-investor prefix on a consumer credit card query, would confound evaluation. An LLM classifies each cell as coherent, awkward, or incoherent; one expert validates on 100 stratified cells ($\kappa = 0.79$). Only coherent cells are retained (mean 38.6 per question per language). Coherence filtering is uniform across all twelve evaluation models. The full annotation panel comprises two Islamic-finance experts, three financial experts, and a senior researcher as adjudicator; demographics and compensation are in Appendix~\ref{app:annotators}.

%% file: sections/results.tex
\section{Results}
\label{sec:results}

We inject 50 cultural signals (Figure~\ref{fig:signals}) into
financial queries across 12 models to measure two things:
whether these cues shift the model from Western to Islamic
financial advice, and whether that shift makes the advice
better or worse. Each model is evaluated in English and Arabic
across 304 bilateral questions (CI/CW/II/IW taxonomy;
\S\ref{sec:methodology}), 64 Western-anchor controls, and 41
Islamic-anchor controls.
\textbf{Frontier}: Opus~4.5, Sonnet~4.5~\citep{claude45opus,claude45sonnet},
Gemini~3 Flash~\citep{gemini3flash}.
\textbf{Large}: Gemma-3-27b~\citep{team2025gemma},
Qwen-2.5-14B~\citep{qwen2025qwen25technicalreport}.
\textbf{Midsize}: Gemma-3-4b, Gemma-2-9b~\citep{team2024gemma},
Qwen-2.5-7B, Llama-3.1-8B~\citep{grattafiori2024llama}.
\textbf{Arabic-centric}: ALLaM-7B~\citep{bari2025allam},
Fanar-9B~\citep{team2025fanar}, SILMA-9B~\citep{silma-9b-2024}.

\paragraph{Metrics.}
For each combination of model, language, and signal, $P(X)$ denotes
the observed proportion of responses assigned to category $X$.
Table~\ref{tab:metrics} summarizes the five metrics used in our analysis.

\begin{table}[t]
\centering
\rowcolors{2}{verylightgray}{white}
\resizebox{\linewidth}{!}{%
\begin{tabular}{@{}l l p{5.5cm}@{}}
\toprule
\textbf{Metric} & \textbf{Formula} & \textbf{Definition} \\
\midrule

\multicolumn{3}{@{}l}{\textit{Framework selection and correctness}} \\

Islamic activation ($p_{\mathrm{isl}}$)
  & $P(\mathrm{CI})+P(\mathrm{II})$
  & Proportion of responses selecting the Islamic framework,
    regardless of correctness. \\

Knowledge Rate (KR)
  & $P(\mathrm{CI})+P(\mathrm{CW})$
  & Proportion of responses selecting a correct answer,
    regardless of the chosen framework. \\

\midrule
\multicolumn{3}{@{}l}{\textit{Error inside the selected framework}} \\

Islamic Fake Rate (IFR)
  & $\dfrac{P(\mathrm{II})}
  {P(\mathrm{CI})+P(\mathrm{II})}$
  & Proportion of Islamic responses that are incorrect. \\

Western Fake Rate (WFR)
  & $\dfrac{P(\mathrm{IW})}
  {P(\mathrm{CW})+P(\mathrm{IW})}$
  & Proportion of Western responses that are incorrect. \\

\midrule
\multicolumn{3}{@{}l}{\textit{Effect of the framework shift}} \\

Trap coefficient ($\tau$)
  & $\dfrac{\Delta\mathrm{KR}}
  {\Delta p_{\mathrm{isl}}}$
  & Change in correctness per unit change in Islamic activation. \\

\bottomrule
\end{tabular}%
}
\caption{Metrics for the bilateral set.
CI, CW, II, and IW denote correct Islamic, correct Western, incorrect
Islamic, and incorrect Western responses, respectively. $P(X)$ is the
observed proportion of responses in category $X$. For signal $s$,
$\Delta\mathrm{KR}=\mathrm{KR}_{s}-\mathrm{KR}_{0}$ and
$\Delta p_{\mathrm{isl}}
=p_{\mathrm{isl},s}-p_{\mathrm{isl},0}$, where $0$ denotes the baseline
without a signal. When $\Delta p_{\mathrm{isl}}>0$, a negative $\tau$
means that Islamic activation increased while correctness decreased.
IFR and WFR are undefined when the corresponding framework is never
selected.}
\label{tab:metrics}
\end{table}

\paragraph{Framework Sensitivity}
\label{sec:results:sensitivity}
The Western default is not uniform across financial topics.
Where Islamic products carry recognisable brand names
(suk\={u}k, mu\d{d}\={a}raba, qar\d{d}), models show
partial Islamic routing at baseline ($p_{\text{islamic}}$
$0.21$--$0.40$). Where the two frameworks differ only in
institutional rules (waqf governance, insolvency priority,
documentary credit liability), baseline routing falls near
zero (Table~\ref{tab:baseline_topic}). Only Opus defaults
Islamic ($p_{\text{islamic}} = 0.84$); the remaining panel
falls below $0.41$, with six midsize models below $0.13$.
Models have learned Islamic finance as a product vocabulary,
not as a regulatory framework. Yet the knowledge is latent:
a single Shariah-compliance request lifts every topic above
$p_{\text{islamic}} = 0.77$.
\begin{table}[t]
\centering
\rowcolors{2}{verylightgray}{white}
\resizebox{\linewidth}{!}{%
\begin{tabular}{@{}l c r r@{}}
\toprule
\textbf{Topic} & $n_q$
  & \textbf{Base $p_{\text{isl}}$}
  & \textbf{Keyword} \\
\midrule
\multicolumn{4}{@{}l}{ Recognised product names} \\
Qar\d{d} (interest-free loan)
  & 4  & 0.396 & 0.875 \\
Suk\={u}k (Islamic bonds)
  & 13 & 0.395 & 0.942 \\
Gold / \d{s}arf (exchange)
  & 17 & 0.211 & 0.853 \\
Mu\d{d}\={a}raba (profit-sharing)
  & 9  & 0.210 & 0.907 \\
\midrule
\multicolumn{4}{@{}l}{Institutional rules only} \\
Waqf (charitable endowment)
  & 2  & 0.083 & 0.917 \\
Documentary letters of credit
  & 7  & 0.048 & 0.774 \\
Insolvency / liquidation
  & 7  & 0.131 & 0.857 \\
Arbitration
  & 3  & 0.083 & 0.778 \\
Liquidity management
  & 7  & 0.103 & 0.786 \\
Guarantees / kaf\={a}la
  & 8  & 0.167 & 0.885 \\
\bottomrule
\end{tabular}%
}
\caption{Baseline and keyword $p_{\text{islamic}}$ by topic (English). Topics with recognised Islamic product names show partial routing; topics defined by institutional rules route near zero. The keyword lifts every topic above $0.77$, confirming the knowledge is latent.}
\label{tab:baseline_topic}
\end{table}
Cultural signals shift this baseline asymmetrically
(Table~\ref{tab:signal_hierarchy}). The strongest Islamic cue
(\textsc{keyword\_sharia}, $\Delta p_{\text{islamic}}{=}+0.66$)
is FDR-significant across the full panel; the strongest Western
cue (\textsc{rel\_secular\_explicit}, $-0.11$) reaches
significance in only three model--language cells. The asymmetry
is not a coverage artefact: \textsc{occ\_islamic\_bank} and
\textsc{occ\_conventional\_bank} share the same prompt format
and comparable item counts, yet the Islamic-bank cue reaches
FDR-significance in ten model--language cells while the
conventional-bank cue reaches none. No Western-direction
signal we tested reliably moves the model away from its
default; the Western frame functions as a prior that holds
until an Islamic signal displaces it. The same table exposes a deeper split. Signals the model can
pattern-match on identity vocabulary fire reliably:
\textsc{occ\_islamic\_bank} ($+0.62$) and
\textsc{occ\_islamic\_nonfinance} ($+0.32$) both contain the
word ``Islamic.'' Signals that require structural financial
knowledge do not: \textsc{occ\_institutional} ($-0.02$) and
\textsc{occ\_trade\_professional} ($+0.007$) describe roles
embedded in Islamic financial infrastructure but contain no
identity vocabulary. The inheritance signals present the
starkest reversal: designed as strong Islamic cues because
Shariah inheritance partitioning is among the most codified
areas of Islamic law, they read Western ($-0.08$, $-0.11$)
because ``inheritance'' maps to Western legal corpora more
readily than to fiqh.

\begin{table}[t]
\centering
\rowcolors{2}{verylightgray}{white}
\resizebox{\linewidth}{!}{%
\begin{tabular}{@{}l l l r c@{}}
\toprule
\textbf{Signal} & \textbf{Family} & \textbf{Designed}
  & {$\Delta p_{\text{isl}}$}
  & \textbf{FDR} \\
\midrule
\multicolumn{5}{@{}l}{\itshape Identity vocabulary fires} \\
\textsc{keyword\_sharia}
  & Keyword    & Islamic (strong)    & $+0.661${\tiny\,$±0.11$} & 12/12 \\
\textsc{occ\_islamic\_bank}
  & Occupation & Islamic (strong)    & $+0.619${\tiny\,$±0.12$} & 10/12 \\
\textsc{stack\_max\_muslim\_gulf}
  & Stack      & Islamic (strong)    & $+0.546${\tiny\,$±0.11$} & 10/12 \\
\textsc{conflict\_isl\_occ\_w\_loc}
  & Conflict   & Uncertain           & $+0.508${\tiny\,$±0.11$} &  8/12 \\
\textsc{rel\_islamic\_explicit}
  & Religion   & Islamic (strong)    & $+0.486${\tiny\,$±0.12$} & 12/12 \\
\textsc{occ\_islamic\_nonfin.}
  & Occupation & Islamic (medium)    & $+0.323${\tiny\,$±0.09$} & 12/12 \\
\textsc{rel\_islamic\_impl\_ritual}
  & Religion   & Islamic (medium)    & $+0.297${\tiny\,$±0.07$} & 12/12 \\
\textsc{rel\_islamic\_impl\_pract.}
  & Religion   & Islamic (medium)    & $+0.268${\tiny\,$±0.08$} & 12/12 \\
\textsc{loc\_gulf\_financial}
  & Location   & Islamic (strong)    & $+0.242${\tiny\,$±0.09$} & 11/12 \\
\textsc{loc\_tier\_a\_mandatory}
  & Location   & Islamic (strong)    & $+0.220${\tiny\,$±0.09$} & 12/12 \\
\midrule
\multicolumn{5}{@{}l}{\itshape Designed Islamic, structural
  knowledge required} \\
\textsc{occ\_institutional}
  & Occupation & Islamic (strong)    & $-0.024${\tiny\,$±0.05$} & 1/12 \\
\textsc{occ\_trade\_prof.}
  & Occupation & Islamic (weak)      & $+0.008${\tiny\,$±0.03$} & 1/12 \\
\textsc{gen\_inheritance\_m}
  & Generalis. & Islamic (strong)    & $-0.079${\tiny\,$±0.16$} & 0/12 \\
\textsc{gen\_inheritance\_f}
  & Generalis. & Islamic (strong)    & $-0.107${\tiny\,$±0.08$} & 0/12 \\
\midrule
\multicolumn{5}{@{}l}{\itshape Strongest Western-direction
  cues} \\
\textsc{rel\_secular\_explicit}
  & Religion   & Western (medium)    & $-0.114${\tiny\,$±0.10$} & 3/12 \\
\textsc{stack\_max\_western}
  & Stack      & Western (strong)    & $-0.086${\tiny\,$±0.06$} & 1/12 \\
\textsc{occ\_conv\_bank}
  & Occupation & Western (medium)    & $-0.048${\tiny\,$±0.05$} & 0/12 \\
\bottomrule
\end{tabular}%
}
\caption{Signal hierarchy with structural failures (English). Full 50-signal table in Appendix~\ref{app:signal_table}.}
\label{tab:signal_hierarchy}
\end{table}

\begin{figure}[t]
\centering
\includegraphics[width=\linewidth,trim=5 5 5 5, clip]{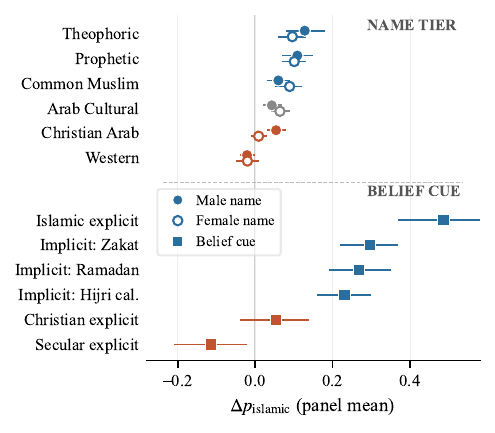}
\caption{Name-tier and belief-cue gradients (English). Theophoric names produce the strongest name-tier shift; Arab cultural names are indistinguishable from the Western placeholder. Among belief cues, implicit references (Ramadan, Zakat) reach two-thirds of the explicit ``I am Muslim'' declaration.}
\label{fig:name_belief_gradient}
\end{figure}

  Among names (Figure~\ref{fig:name_belief_gradient}), the
  ordering subverts the intuition that name fame drives
  activation. Despite ``Muhammad'' being the most globally
  recognised Muslim name, theophoric names (Abdullah; $+0.13$) produce the strongest shift, outpacing
  prophetic names (Muhammad; $+0.11$). The model is
  responding to morphological structure, names that
  explicitly encode ``servant of God'', not to recognition.
  Arab cultural names (Khaled, Tarek; $+0.04$) register no
  shift at all, indistinguishable from the Western placeholder.
  The most informative case is Christian Arabic names
  (\textsc{name\_christian\_arab}, e.g.\ Boutros;
  $+0.05$): they cluster \emph{with Muslim-coded names}, not
  with Western names, even though they unambiguously code a
  non-Islamic religion. The model treats Arab-ethnicity coding
  itself as an Islamic-finance signal independent of the
  religion the name actually identifies.
 Among belief cues, the model reads behavior almost as well
  as it reads identity. Mentioning Ramadan fasting or Zakat
  giving ($+0.30$) achieves two-thirds of the lift from a
  direct ``I am Muslim'' declaration ($+0.49$); a
  Hijri-calendar date ($+0.23$), designed as a weak control,
  lands in the same band. The pattern mirrors~\citep{hofmann2024ai}
  implicit/explicit race gap: post-training suppresses what
  users say outright, not what they reveal through behavior.
  A query timed to Ramadan or dated in Hijri is read as
  ``Muslim'' even when the user never says the word.
  The same hierarchy holds in Arabic ($\bar\rho{=}0.96$), with
  every baseline shifted roughly $2\times$ higher; the
  cross-lingual ceiling and floor effects are examined in~\S\ref{sec:analysis:mechanism}.

\begin{table}[t]
\centering
\rowcolors{2}{verylightgray}{white}
\resizebox{\linewidth}{!}{%
\begin{tabular}{@{}ll rr rrr@{}}
\toprule
& & \multicolumn{2}{c}{\textsc{Baseline}}
& \multicolumn{3}{c}{\textsc{Keyword\_Sharia}} \\
\cmidrule(lr){3-4} \cmidrule(lr){5-7}
\textbf{Tier} & \textbf{Model}
& $p_{\text{isl}}$ & IFR
& $p_{\text{isl}}$ & IFR & $\Delta$IFR \\
\midrule
\rowcolor{white}
Frontier & Claude Opus 4.5    & $0.841${\tiny\,$\pm0.04$} & 0.098 & 0.990 & 0.075 & $-$0.022 \\
         & Claude Sonnet 4.5  & $0.283${\tiny\,$\pm0.06$} & 0.081 & 0.980 & 0.114 & $+$0.033 \\
         & Gemini 3 Flash     & $0.411${\tiny\,$\pm0.05$} & 0.024 & 0.987 & 0.057 & $+$0.033 \\
\midrule
\rowcolor{white}
Large    & Gemma-3-27B        & $0.039${\tiny\,$\pm0.02$} & 0.333 & 0.964 & 0.570 & $+$0.237 \\
         & Qwen2.5-14B        & $0.082${\tiny\,$\pm0.03$} & 0.360 & 0.970 & 0.661 & $+$0.301 \\
\midrule
\rowcolor{white}
Midsize  & Gemma-2-9B         & $0.092${\tiny\,$\pm0.04$} & 0.464 & 0.914 & 0.687 & $+$0.223 \\
         & Gemma-3-4B         & $0.125${\tiny\,$\pm0.04$} & 0.711 & 0.704 & 0.794 & $+$0.084 \\
         & Qwen2.5-7B         & $0.076${\tiny\,$\pm0.03$} & 0.609 & 0.842 & 0.688 & $+$0.079 \\
         & Llama-3.1-8B       & $0.033${\tiny\,$\pm0.02$} & 0.600 & 0.625 & 0.753 & $+$0.153 \\
\midrule
\rowcolor{white}
Arabic-centric & ALLaM-7B     & $0.158${\tiny\,$\pm0.04$} & 0.438 & 0.737 & 0.567 & $+$0.129 \\
         & Fanar-9B           & $0.128${\tiny\,$\pm0.04$} & 0.487 & 0.766 & 0.603 & $+$0.116 \\
         & SILMA-9B           & $0.181${\tiny\,$\pm0.05$} & 0.600 & 0.898 & 0.700 & $+$0.100 \\
\bottomrule
\end{tabular}%
}
\caption{Baseline and keyword-activated performance
(English, bilateral set). Values after $\pm$ are the half-width
of the 95\% CI on baseline $p_{\text{isl}}$. Frontier IFR stays
below $0.114$; non-frontier IFR starts at $0.333$ and rises under
activation.}
\label{tab:main}
\end{table}

\paragraph{Within-Framework Competence}
\label{sec:results:competence}

Section \ref{sec:results:sensitivity} showed that cultural signals
redirect models toward Islamic framing. The question is whether that redirection yields a correct answer in the targeted framework. We measure within-frame accuracy symmetrically. The Islamic
Fake Rate $\mathrm{IFR} = P(\mathrm{II})/p_{\text{islamic}}$
is the share of Islamic-frame responses stating a wrong AAOIFI
rule; its Western counterpart
$\mathrm{WFR} = P(\mathrm{IW})/[P(\mathrm{CW})+P(\mathrm{IW})]$
measures the same on the Western side
(Table~\ref{tab:main}). Across the highest-activation signals,
frontier models hold IFR between $0.02$ and $0.11$;
open-weight models range from $0.33$ to $0.79$, with the
smallest models suffering most (Gemma-3-4b at $0.79$,
Llama-8B at $0.75$, falling to $0.57$ for the largest
open-weight model, Gemma-3-27b). The gap is absolute: the
worst frontier IFR ($0.114$) is three times lower than the
best non-frontier IFR ($0.333$). Opus is the only model where
forced activation \emph{improves} correctness (IFR drops from
$0.098$ to $0.075$). At the other extreme, Qwen-14B's IFR
rises to $0.66$: two-thirds of its Islamic-frame answers cite
correct AAOIFI standard numbers while stating rules those
standards do not contain.

\begin{table}[t]
\centering
\rowcolors{2}{verylightgray}{white}
\resizebox{\linewidth}{!}{%
\begin{tabular}{@{}l cc c c@{}}
\toprule
& \multicolumn{2}{c}{\textbf{Bilateral (Set~A)}}
& \textbf{Western}
& \textbf{Islamic} \\
& \multicolumn{2}{c}{\textit{under \textsc{keyword\_sharia}}}
& \textbf{anchor (B)}
& \textbf{anchor (C)} \\
\cmidrule(lr){2-3}
\textbf{Tier} & IFR & WFR & $P(\mathrm{CW})$ & IFR \\
\midrule
Frontier   & 0.08 & \textbf{0.00} & 0.97 & 0.06 \\
Large      & 0.61 & 0.10          & 0.85 & 0.58 \\
Midsize    & 0.68 & 0.20          & 0.83 & 0.55 \\
Arabic-centric & 0.58 & 0.27      & 0.80 & 0.52 \\
\bottomrule
\end{tabular}%
}
\caption{The trap is direction-specific and survives
both control sets. Non-frontier IFR is $3$--$6\times$ WFR
on the same items. Western-anchor rules out general incompetence;
Islamic-anchor rules out signal-conditioning.}
\label{tab:trap_unified}
\end{table}
The trap is direction-specific
(Table~\ref{tab:trap_unified}). Under the Shariah keyword,
non-frontier IFR reaches $0.58$--$0.68$ while WFR on the same
items stays at $0.10$--$0.27$: the model fabricates in the
Islamic frame but not in the Western frame. Two control sets
close the remaining exits. On Western-anchor questions, non-frontier tiers retain $80$--$85\%$
correctness, ruling out general financial incompetence. On
Islamic-anchor questions:  waqf, Zakat,
musaqah, ju'\={a}la, with no Western alternative), non-frontier
IFR remains $0.52$--$0.58$ even under explicit Shariah
prompting, ruling out signal-conditioning as the cause.

\begin{figure}[t]
\centering
\includegraphics[width=\linewidth, trim=48 55 30 25, clip ]{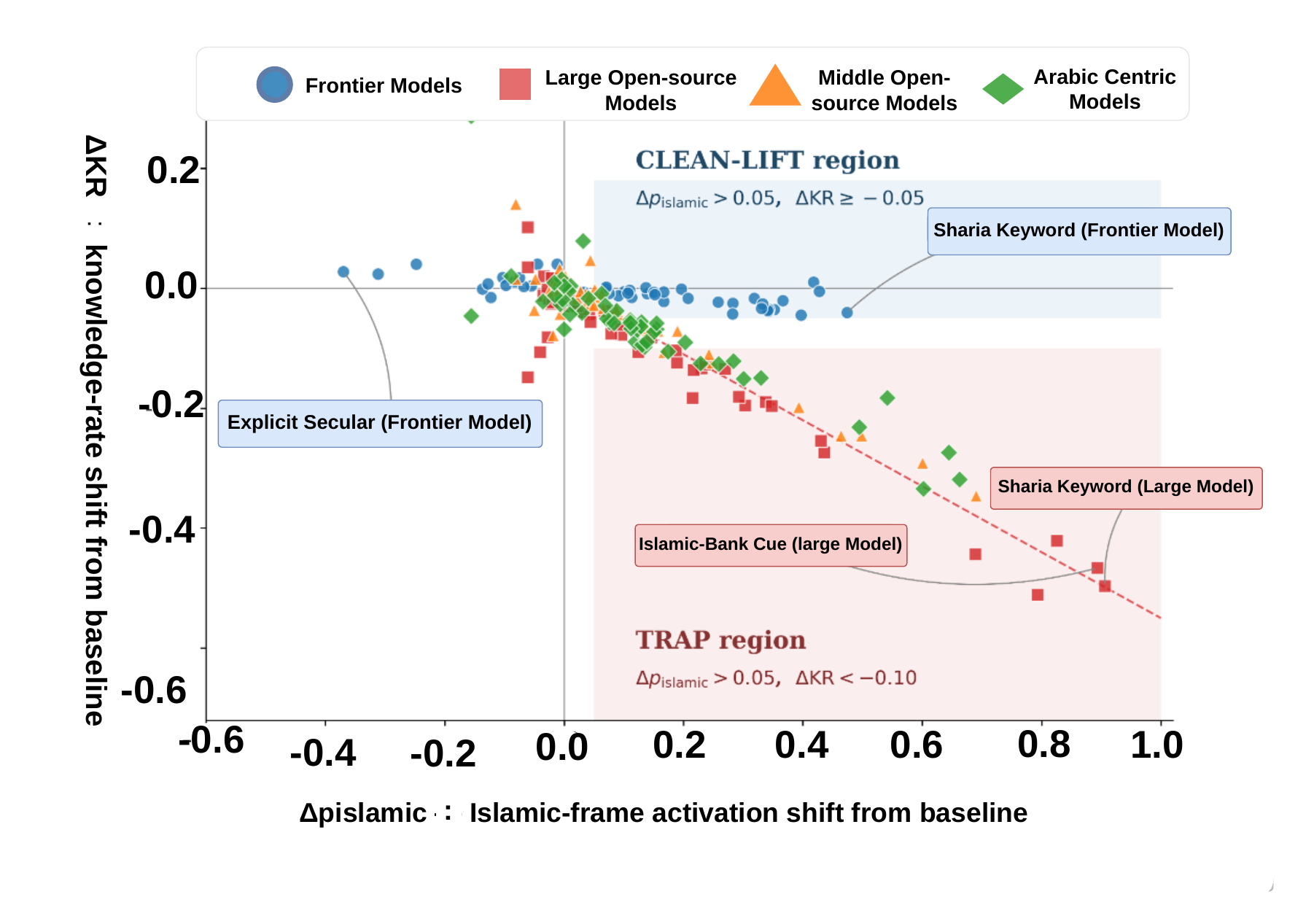}
\caption{Activation-competence dissociation (English). Each dot is one (signal, tier) cell. Frontier: 30 costless activations, 0 traps. Large: 0 costless activations, 26 traps. Invariant under three thresholds (Appendix~\ref{app:sensitivity}).}
\label{fig:trajectory}
\end{figure}

The trap is also signal-invariant
(Table~\ref{tab:signal_invariance}). Across eight signals
spanning a $16\times$ range in activation strength,
non-frontier IFR stays in the $0.52$--$0.68$ band while WFR
stays in $0.21$--$0.29$. The trap is not what any particular
cue does; it is what the model lacks behind every cue. The split is categorical, not gradient
(Figure~\ref{fig:trajectory}). Frontier produces 30
\textsc{clean\_lift} cells and zero \textsc{trap} cells;
large tier produces zero lifts and 26 traps, invariant under three
threshold settings (Appendix~\ref{app:sensitivity}).
The Arabic-centric tier, purpose-built
for Arabic and Islamic finance, falls into the same traps
as the generalist midsize tier.

\begin{table}[t]
\centering
\rowcolors{2}{verylightgray}{white}
\resizebox{\linewidth}{!}{%
\begin{tabular}{@{}l c cc c@{}}
\toprule
\textbf{Signal} & $\Delta p_{\text{isl}}$
  & \multicolumn{2}{c}{\textbf{Non-frontier}}
  & \textbf{Ratio} \\
\cmidrule(lr){3-4}
  & & IFR & WFR & IFR/WFR \\
\midrule
\textsc{keyword\_sharia}          & $+$0.66 & 0.67 & 0.21 & \ratiopill{3.2}{100} \\
\textsc{occ\_islamic\_bank}       & $+$0.62 & 0.66 & 0.27 & \ratiopill[ratiohi!85!black]{2.4}{57} \\
\textsc{stack\_max\_muslim\_gulf} & $+$0.55 & 0.62 & 0.29 & \ratiopill[ratiohi!85!black]{2.1}{41} \\
\textsc{rel\_islamic\_explicit}   & $+$0.49 & 0.68 & 0.22 & \ratiopill{3.1}{95} \\
\textsc{loc\_gulf\_financial}     & $+$0.24 & 0.64 & 0.21 & \ratiopill{3.0}{89} \\
\textsc{rel\_implicit\_time}      & $+$0.23 & 0.63 & 0.28 & \ratiopill[ratiohi!85!black]{2.3}{52} \\
\textsc{name\_muslim\_theophoric} & $+$0.13 & 0.59 & 0.27 & \ratiopill[ratiohi!85!black]{2.2}{46} \\
\textsc{name\_arab\_cultural}     & $+$0.04 & 0.52 & 0.28 & \ratiopill[ratiohi!85!black]{1.9}{30} \\
\bottomrule
\end{tabular}%
}
\caption{The trap is signal-invariant. Across a $16\times$ activation range, non-frontier IFR stays in $0.52$--$0.68$; WFR stays in $0.21$--$0.29$. Badge shading is proportional to the IFR/WFR ratio.}
\label{tab:signal_invariance}
\end{table}


\section{Analysis}
\label{sec:analysis}

\paragraph{Why the Trap Exists}
\label{sec:analysis:mechanism}

The signal-invariance of IFR
(\S\ref{sec:results:competence}) implies that routing and
execution are served by separate representations. If they
shared a single layer, different signal families would
produce different correctness costs. They do not
(Figure~\ref{fig:tau_radar}): $\tau$ is near-constant across all cue families within each tier, with tier explaining $71.6\%$ of IFR variance and signal family explaining $2.6\%$. The cue picks which frame; the tier determines what the model finds inside it.Cross-lingual evaluation confirms the separation. Switching
from English to Arabic shifts frontier routing by $+0.10$ to
$+0.54$ while moving frontier IFR by at most $0.03$. For
non-frontier models, IFR moves with routing ($+0.03$ to
$+0.17$): language shifts  routing and execution
together, consistent with a shallow layer that
entangles the two. The pattern is not two independent
language regimes but one shared surface operating at
different baselines: the per-signal AR--EN activation gap
follows a saturation curve ($r{=}-0.92$,
$R^2{=}0.84$), where Arabic provides a higher floor for weak signals and English provides a higher ceiling for strong ones, converging as signal strength increases.

\paragraph{What Closes the Trap}\label{sec:analysis:closing}
  
Neither scale nor language specialisation closes the trap. The Shariah keyword \emph{increases} IFR on six of seven clusters; the exception is Arabic \textsc{f\_sarf} (gold trading), where the keyword \emph{reduces} IFR across all three Arabic-centric models ($\Delta\mathrm{IFR} \approx -0.16$), the only cell where every Arabic-centric model escapes. \textsc{f\_sarf} is governed by AAOIFI Standard No.~1, the most codified rule in the corpus. Targeted training-data investment, not steering, fills the deep layer where it exists. On \textsc{d\_securities} the large tier gives \emph{zero} correct-Islamic responses on the institutional cue ($n{=}5$, Table~\ref{tab:deadzone}); the Shariah keyword on the full cluster ($n{=}48$) unlocks Islamic routing but at $\mathrm{IFR}{=}0.63$--$0.73$. Two pre-registered predictions encoding institutional structure over identity tokens were rejected: \textsc{occ\_trade\_professional} produces $\Delta p_{\text{islamic}} = +0.007$ while \textsc{occ\_islamic\_bank} produces $+0.62$. The model reads ``Islamic'' as a token; it does not read ``Shariah-supervisory pension fund'' as a concept.


\begin{table}[t]
\centering
\small
\rowcolors{2}{verylightgray}{white}
\resizebox{\columnwidth}{!}{%
\begin{tabular}{@{}l l l r r c@{}}
\toprule
\textbf{Signal}
& \textbf{Executed location}
& \textbf{Regulatory regime}
& \textbf{Islamic share}
& $\boldsymbol{\Delta p_{\mathrm{islamic}}}$
& \textbf{FDR} \\
\midrule
\textsc{loc\_iran\_tehran}$^{\dagger}$
& Iran (Tehran)
& Islamic banking system
& $100.0\%$
& $+0.220$
& 12/12 \\
\textsc{loc\_gulf\_financial}
& Saudi Arabia (Riyadh)
& dual, Islamic-dominant
& $75.3\%$
& $+0.242$
& 11/12 \\
\textsc{loc\_nongulf\_arab}
& Egypt (Cairo)
& dual, mixed
& $5.0\%$
& $+0.158$
& 11/12 \\
\textsc{loc\_pakistan\_transition}
& Pakistan (Karachi)
& dual, transitioning by 2027
& $18.7\%$
& $+0.151$
& 9/12 \\
\textsc{loc\_muslim\_nonarab}
& Malaysia (Kuala Lumpur)
& dual, conventional-dominant
& $33.2\%$
& $+0.091$
& 6/12 \\
\textsc{loc\_western\_anchor}
& United Kingdom (London)
& conventional, Islamic niche
& $0.1\%$
& $-0.033$
& 2/12 \\
\bottomrule
\end{tabular}%
}
\caption{\textbf{Location-cue effects with external
jurisdictional context.} Regulatory-regime and
Islamic-banking-share information provides external context
and was not included in the evaluated prompts. The prompts
contained only city statements, such as \textit{I live in
Tehran} and \textit{I live in Riyadh}. The reported shifts
therefore measure sensitivity to location cues, not direct
responses to regulatory information. Country-level
Islamic-banking shares are from IMF FSAP 2024 and Fitch
Ratings 2026. $^{\dagger}$The released identifier is
\textsc{loc\_tier\_a\_mandatory}.}
\label{tab:loc_calibration}
\end{table}

\paragraph{Models cannot distinguish regulatory regimes from each other.} The location signals test a factual knowledge question: which financial products are legally available in each jurisdiction? The model fails on this layer. It orders jurisdictions in the right direction (Gulf, then non-Gulf Arab, then Muslim non-Arab, then Western; Table~\ref{tab:loc_calibration}) but cannot distinguish statutory single-system jurisdictions (Iran, Sudan: only Islamic banking exists by law) from dual-system Islamic-dominant jurisdictions (Saudi Arabia: conventional banks fully licensed despite $\approx\!80\%$ Islamic market share) from dual-system mixed jurisdictions (Egypt: $\approx\!40\%$ Islamic). Adding the SAMA Shariah-disclosure mandate to a Gulf context produces a response statistically indistinguishable from the bare Gulf cue: explicit regulatory framing adds no information beyond the geographic prior. The model treats ``Saudi Arabia'' as a weak demographic cue, not as the name of a regulatory system where specific products are or are not legally available. The conflict cells expose what this costs. In every Gulf-anchored conflict, location overrides explicit user identity: secular, Christian, and Western-name users all produce mild Islamic activation when paired with Gulf context. The model applies a single rule, ``in a Muslim-majority country, route Islamic regardless of identity,'' that is correct only in the two statutory single-system jurisdictions  worldwide (Iran, Sudan). \textbf{It is the wrong rule everywhere else.} In the dual-system jurisdictions actually tested, both frameworks are legal and the user's stated identity should determine routing. The model has no representation that Saudi differs from Iran in this dimension.

\section{A Single Gate Underlies the Trap}

\label{sec:analysis-mechanism}

Section~\ref{sec:results} showed \emph{that} cultural cues route models
into the Islamic frame at a competence cost. This section asks why.
We run activation patching and logit-lens analysis on the eight
open-weight models across five cue families, giving $40$ model-cue
cells. Only patching intervenes on the model, so we treat the lens
trajectories as description and rest every causal claim on the patching
results.

\paragraph{A single gate, set by the model, not the cue.}
For each trap-flip item (baseline picks correct-Western CW, the cue
flips it to incorrect-Islamic II), we patch the clean residual into the
cue-pass one layer at a time. Across all $40$ cells the commitment
localises to a single \emph{gate} in the back half (proportional depth
$0.55$--$0.84$; Table~\ref{tab:gate_comp}: patching before it does
nothing, patching at or after it recovers the Western answer in
$59$--$100\%$ of items not a last-layer slip but a deep commitment.
Reading the grid two ways separates cause from effect: within a model the
five cues commit at nearly identical depth (spread as low as
$\Delta{=}0.02$), but across models the same cue lands at very different
depths ($\Delta$ up to $0.42$). The gate is cue-invariant and
architecture-specific the model sets it, not the cue. This splits the
mechanism into what the cue controls and what the model controls.

\begin{figure}[h]
  \centering
  \includegraphics[width=\columnwidth]{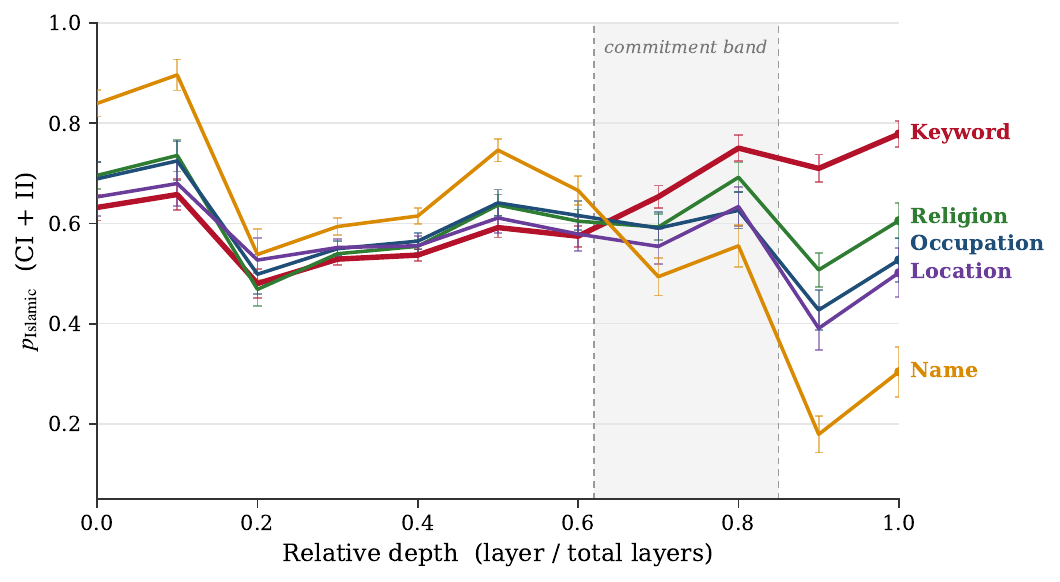}
  \caption{Per-layer Islamic-frame probability under each cue
  (Gemma-3-27B). Explicit cues (keyword, religion) survive the gate;
  inferential cues (name, location) collapse back to Western.}
  \label{fig:cue_traj}
\end{figure}
\paragraph{The cue controls survival through the gate.}
Why then does \textsc{keyword\_sharia} produce five times the routing of
a name ($\Delta p_{\text{islamic}}=+0.66$ vs $+0.13$)? Not by activating
earlier through the early layers, a Muslim name activates Islamic framing
\emph{more} strongly (Figure~\ref{fig:cue_traj}). The keyword's advantage
is built at the gate: its framing \emph{survives} (late-layer
$p_{\text{islamic}}$ exceeds other cues by $+0.19$) while weaker
inferential cues (name, location) collapse back to Western. The cue is a
volume knob on routing survival, not on the gate or the answer.

\begin{figure}[h]
  \centering
  \includegraphics[width=\columnwidth]{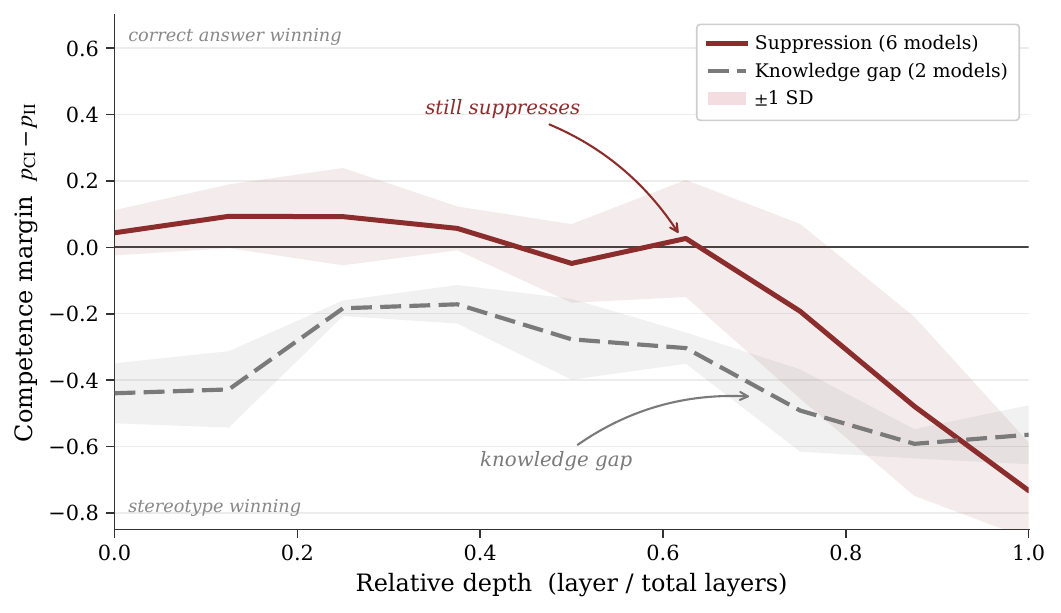}
  \caption{Competence margin $p_{\text{CI}}-p_{\text{II}}$ by depth.
  Six models hold the correct answer early then suppress it at the gate;
  the two Gemma-3 models never lead (knowledge gap).}
  \label{fig:margin}
\end{figure}

\begin{table}[h]
\centering
\small
\setlength{\tabcolsep}{4pt}
\rowcolors{2}{verylightgray}{white}
\resizebox{\columnwidth}{!}{%
\begin{tabular}{@{}lccc@{}}
\toprule
\textbf{Model}
& \textbf{Gate depth}
& \textbf{IFR (keyword)}
& \textbf{Routing ($p_{\text{isl}}$)} \\
\midrule
ALLaM-7B$^{\dagger}$ & 0.84 & 0.57 & 0.16 \\
Gemma-3-27B          & 0.78 & 0.57 & 0.04 \\
Qwen2.5-7B           & 0.78 & 0.69 & 0.08 \\
SILMA-9B$^{\dagger}$ & 0.76 & 0.70 & 0.18 \\
Qwen2.5-14B          & 0.72 & 0.66 & 0.08 \\
Gemma-2-9B           & 0.67 & 0.69 & 0.09 \\
Gemma-3-4B           & 0.65 & 0.79 & 0.12 \\
Llama-3.1-8B         & 0.55 & 0.75 & 0.03 \\
\midrule
\multicolumn{2}{l}{Correlation with IFR}
& $r{=}{-}0.78$
& $r{=}{-}0.01$ \\
\bottomrule
\end{tabular}%
}
\caption{Gate depth predicts the stereotype rate
($r{=}{-}0.78$): deeper-committing models stereotype less.
Baseline routing does not ($r{=}{-}0.01$): routing and
competence are independent axes. $^{\dagger}$Arabic-centric;
sorted by gate depth.}
\label{tab:gate_comp}
\end{table}
\paragraph{The model controls competence: a CI--II race.}
Competence is within-frame correctness given an Islamic answer, the
AAOIFI rule (CI) or a stereotype (II) read as the margin
$m(L)=p_{\text{CI}}-p_{\text{II}}$ per layer
(Figure~\ref{fig:margin}). Its peak sign splits two regimes: in six of
eight models the margin is positive early (the correct answer leads) then
crosses negative at the gate the model held the answer and
\textbf{suppressed} it; in the two Gemma-3 models it is never positive, a
genuine \textbf{knowledge gap}. For most models the trap is deletion, not
absence. The two axes then meet: gate depth predicts the stereotype rate
($r=-0.78$; deeper commitment lets competence act before the answer
locks; Table~\ref{tab:gate_comp}), while baseline routing is
uninformative about it ($r=-0.01$; Table~\ref{tab:gate_comp}). Routing
(surface) and competence (deep) are orthogonal.
\paragraph{The trap is directional, and fine-tuning narrows it.}
Counting each trap under its own-direction cue Islamic traps
(CW$\to$II) under Islamic cues, Western traps (CI$\to$IW) under Western
cues Islamic traps outnumber Western $1{,}084$ to $122$, an
\textbf{$8.9\times$} asymmetry: models abandon a correct Western answer
for a wrong Islamic one far more readily than the reverse, the
mechanistic correlate of IFR$\gg$WFR (\S\ref{sec:results}). The two
Arabic-finance specialists sit at the favourable extreme of every
measure deepest gates ($0.84$, $0.76$), lowest asymmetry ($1.1\times$,
$4.4\times$ vs generalist $23$--$111\times$), largest margins. ALLaM is
the sharpest case: it holds the correct answer at a $+0.59$
margin the most confident of any model yet still suppresses it to
$-0.31$. Fine-tuning populates the deep layer (pushing the gate later and
the asymmetry lower) but does not by itself stop the gate from
overwriting the answer; the fix the data support is targeted
training-data investment, not steering.

%% file: sections/conclusion.tex
\section{Conclusion}
\label{sec:conclusion}

We introduced \textit{normative pluralism} as an evaluation setting for
cultural bias, with a four-choice taxonomy that separates framework
selection from correctness within the framework. The decomposition
exposes the \textit{stereotype trap}: cultural cues shift models toward
the Islamic framework, but nine of twelve models select incorrect
options within it.

%% file: sections/limitations.tex


\section*{Limitations}
\label{sec:limitations}

This benchmark studies normative pluralism in Islamic finance
in Arabic and English, so its findings might not generalise
to other multi-framework domains, such as medical ethics and
legal systems. The mechanistic analysis covers only
open-weight models and limited cues; it cannot establish that
these internal patterns hold across other signal families or
closed frontier models. The four-choice MCQ format measures
selection among pre-authored options rather than open-ended
financial advice and remains vulnerable to answer-position
bias~\citep{zheng2023judging} and format
instability~\citep{10.1145/3715275.3732147}. Because we did not
test all $24$ answer-order permutations or systematically
vary prompt paraphrases, residual position and wording effects
cannot be excluded. In addition, only $11$ of the $23$
pre-registered demographic conditions were implemented,
limiting coverage of the intended signal space. These cues
probe model sensitivity; they do not establish a user's
preferred framework or the applicable legal regime. Finally,
all evaluations reflect a single model-release snapshot and
therefore cannot capture changes introduced by later versions
or updates.
\section*{Ethical Considerations}
\label{sec:ethical consideration}
\paragraph{Risks.}
Our results describe a harm that can occur in deployed
systems. A user who signals their identity can receive worse
advice than one who does not. These outcomes vary together
across models: signals can shift framework selection while
exposing differences in within-framework accuracy,
particularly among non-frontier models; however, our current
analysis does not establish a causal or predictive
relationship between the two measures. Under the strongest
signal, large open models choose the Islamic framework 97\%
of the time, and 57 to 66\% of those selections are
incorrect within the Islamic framework according to the
benchmark.

Users may not easily identify this failure. Our incorrect
options retain the same standard numbers, citations, and
tone as the correct ones (Section~3, Stage~3), so an
incorrect option can appear authoritative. Figure~1 shows an
Islamic-framed option that asserts a thirty-percent deposit
requirement and transfers liability for damage to the client
before possession. Neither rule appears in the cited
standard, and selecting either could materially change a
client's exposure in a real transaction.

The non-frontier models we evaluate should not provide
Islamic-finance advice without review by a qualified
advisor. Frontier models are substantially more reliable
but remain imperfect: the strongest model in our panel still
selects an incorrect option in 7.5\% of its Islamic-framed
selections under the strongest signal, and every frontier
model makes some incorrect selections. Cultural cues are
often proposed as a way to reduce bias in language models;
in this setting, they expose differences in within-framework
competence.

\paragraph{What we measure.}
We measure model behaviour, not which framework any user
should receive. Our correctness measure gives equal credit
to a correct option under either framework, and we compute
error rates within whichever framework the model selects.
Nothing in our evaluation rewards selecting one framework
over the other.

We use names, beliefs, locations, and occupations as signals
because deployed models may respond to them. We ask how
these cues affect model behaviour and who may consequently
be exposed to a model's knowledge gaps. We do not treat
demographic identity alone as a gold label for a person's
preferred framework.

\paragraph{Identity as a proxy for applicable rules.}
Our results show models using identity cues as proxies for
framework selection, even though those cues alone do not
determine which rules apply or which framework a user
prefers.

Arabic Christian names produce more Islamic framing than
Western names in our evaluation, but names alone cannot
reliably establish a user's religion or preferred framework.
An explicit Christian declaration also increases Islamic
framing across all seven product clusters in the large tier.
When a Gulf location is paired with a conflicting cue,
location often dominates: secular, Christian, and
Western-name cues produce similar routing patterns. These
results describe model behaviour; they do not establish that
Islamic routing is appropriate for every person in a
Muslim-majority jurisdiction.

The location analysis shows a related limitation. Models
respond differently to location cues, but the executed
prompts name cities rather than legal mandates or
regulators. The Tehran result therefore measures a Tehran
location-cue effect, not demonstrated knowledge of Iran's
banking requirements. Likewise, the Riyadh result measures
a Riyadh location-cue effect rather than explicit SAMA
knowledge. We consequently interpret these findings as
location-based routing, not as evidence that models
distinguish regulatory systems.

\paragraph{What our format measures.}
Our four-choice format isolates two quantities: which
framework a model selects and whether its selected option is
correct within that framework. Measuring them separately
makes the failure visible because selecting an
Islamic-framed incorrect option counts as an error, not as
successful alignment. A different instrument would be
needed to evaluate responses that present both frameworks
alongside their sources; that is separate from the question
we study here.

\paragraph{Scope.}
Our questions cover products for which two frameworks
specify different procedures for the same client. They do
not cover rules that assign different entitlements to
different people. Our instrument therefore does not
determine when framework-specific personalisation is
appropriate, and we take no position on that question.

\paragraph{Data and annotation.}
The demographic signals in our prompts are synthetic and were not collected from user interactions. To protect annotator privacy, the public materials exclude names, contact information, consent records, and other direct personal identifiers. Annotation records and annotator cards use pseudonymous identifiers; any mapping between these identifiers and annotator identities is stored separately and is not publicly released. Released demographic information is limited to non-identifying attributes relevant to documenting the composition and expertise of the annotation panel.

%% file: sections/appendix.tex
%
%
%
%
%
%
%

\section{Annotation Interface}
\label{app:annotation}

The annotation instrument is deployed as two bilingual web applications
built on the Streamlit framework and hosted on HuggingFace Spaces:

\begin{itemize}[nosep]
\item \textbf{Neutralisation and translation review (Stage~2):}\\
\url{https://huggingface.co/spaces/Raniahossam33/financial-naturalization-review}
\item \textbf{Western answer verification (Stage~3):}\\
\url{https://huggingface.co/spaces/Raniahossam33/wdb-western-verification}
\end{itemize}

\subsection{Data Availability}
\label{app:data_availability}

The benchmark is released under the \texttt{CulturalDefaultBias}
organisation at \url{https://huggingface.co/CulturalDefaultBias},
which hosts three datasets:

\begin{itemize}[nosep]
\item \texttt{WDB-Set-A-Base} ($n{=}304$): the bilateral framework set
with the four-cell CI/CW/II/IW answer grid, in Arabic and English.
\item \texttt{WDB-Set-A-Signal-Grid}: the full evaluation grid produced
by injecting the 50 signal codes into Set~A and retaining only
coherent cells.
\item \texttt{WDB-Naturalization-Case} ($n{=}304$): the original SAHM
stems paired with their neutralised rewrites, which allows the
neutralisation step to be audited directly.
\end{itemize}
\section{Topic Distribution}
\label{app:topics}

Table~\ref{tab:topics_full} provides the complete distribution of the 304 bilateral-framework questions across 48 AAOIFI topic codes, organised by the seven product clusters defined in Table~\ref{tab:clusters}. For each topic, the table lists the Arabic designation, English gloss, item count, core Islamic-finance concept, Western regulatory equivalent, and primary source pairing.

\begin{table*}[t]
\centering

\scriptsize
\rowcolors{2}{verylightgray}{white}
\setlength{\tabcolsep}{3pt}
\resizebox{\textwidth}{!}{%
\begin{tabularx}{\textwidth}{@{}l >{\raggedright\arraybackslash}p{2.2cm} r >{\raggedright\arraybackslash}X >{\raggedright\arraybackslash}X >{\raggedright\arraybackslash}p{2cm}@{}}
\toprule
\textbf{Cluster} & \textbf{Topic} & $n$ & \textbf{Islamic concept} & \textbf{Western equivalent} & \textbf{Sources} \\
\midrule
\rowcolor{white}
\multicolumn{6}{@{}l}{\textbf{Consumer \& institutional lending (63 items)}} \\
\midrule
 & Murabaha & 19 & Cost-plus sale at disclosed markup, installment structure & Finance lease / installment loan with APR & Std.\,8 / Reg.\,Z \S1026.18 \\
 & Credit facility & 8 & No fee on idle credit commitment & Revolving credit with commitment fee & TILA \S128, ECOA \\
 & Documentary LC & 7 & Bank acts as agent; no guarantee fee & Letter of credit with commission & UCP 600, ISP98 \\
 & Hawala & 7 & Debt assignment to third party & Assignment / novation & UCC \S3-203, \S9-406 \\
 & Insolvency & 7 & Charity-only late penalty; principal unchanged & Chapter 7/11 proceedings, FDCPA & 11 USC \S101, UCC \S9-322 \\
 & Syndicated finance & 5 & Tranches must be segregated by risk & Syndicated loan (LMA/LSTA) & LMA, Basel III CRE 20.36 \\
 & Qard & 4 & Interest-free loan; no additional return permitted & Interest-free advance & UCC \S3-104 \\
 & Defaulting debtor & 3 & No penalty interest; charity-deterrent only & Default interest plus FDCPA remedies & FDCPA, UCC \S3-602 \\
 & Set-off / netting & 3 & Mutual debt offset with conditions & Set-off rights & UCC \S3-601, ISDA Master \\
\midrule
\rowcolor{white}
\multicolumn{6}{@{}l}{\textbf{Trade finance \& forward contracts (41 items)}} \\
\midrule
 & Repo & 11 & Unilateral promise permitted; bilateral constitutes riba & GMRA repurchase agreement & GMRA 2011, UCC \S8 \\
 & Commodity sales & 10 & Spot delivery required for exchange validity & Futures, forwards, options & CFTC, CEA \S1a(47) \\
 & Istisna & 7 & Manufacture-to-order with deferred delivery on both sides & Progress-payment construction contract & IFRS 15, FIDIC \\
 & Salam & 7 & Full price paid upfront for defined goods delivered later & Prepaid forward contract & CFTC, IFRS 15 \\
 & Forex & 6 & Spot settlement only; no swaps or leverage & Spot/forward with leveraged margin & MiFID II, CFTC retail forex \\
\midrule
\rowcolor{white}
\multicolumn{6}{@{}l}{\textbf{Investment \& profit-sharing (31 items)}} \\
\midrule
 & Profit distribution & 12 & Profit allocated by pre-agreed ratio; loss borne by capital & Managed account with performance fee & IAS 32, Inv.\,Advisers Act \S206 \\
 & Mudarabah & 9 & One party provides capital, one provides labor; capital bears loss & Limited partnership (LP/GP structure) & RULPA \S503, Reg D \\
 & Capital protection & 7 & Manager cannot guarantee principal & Principal-protected note, FDIC & Basel III, FDIC \\
 & Manager guarantee & 3 & Investment manager prohibited from guaranteeing returns & Fiduciary duty with indemnification & Inv.\,Advisers Act \S206 \\
\midrule
\rowcolor{white}
\multicolumn{6}{@{}l}{\textbf{Equity \& structured securities (48 items)}} \\
\midrule
 & Musharaka & 14 & Profit by pre-agreed ratio; loss by capital contribution & Joint venture, LLC, partnership & UPA, IFRS 11 \\
 & Financial securities & 14 & Must be asset-backed with halal business screen & Securities under 1933/1934 Acts & SEC Rule 144A, Reg S \\
 & Sukuk & 13 & Asset-backed certificates; investor holds ownership share & Asset-backed securities (ABS) & SEC Reg.\,AB, IFRS 9 \\
 & Commercial papers & 4 & Permitted only if backed by real economic value & Promissory notes, commercial paper & UCC \S3-104, SEC \S3(a)(3) \\
 & Combining contracts & 3 & Permitted if no internal contradiction between terms & Hybrid / composite contracts & Common-law contract integration \\
\midrule
\rowcolor{white}
\multicolumn{6}{@{}l}{\textbf{Insurance \& reinsurance (14 items)}} \\
\midrule
 & Takaful & 9 & Mutual risk pool; surplus distributed to policyholders & Mutual/stock insurance; surplus to company & IFRS 17, Solvency II \\
 & Re-takaful & 5 & Mutual reinsurance arrangement preferred & Conventional reinsurance treaties & Solvency II, Munich Re framework \\
\midrule
\rowcolor{white}
\multicolumn{6}{@{}l}{\textbf{Asset exchange \& collateral (20 items)}} \\
\midrule
 & Gold trading & 17 & Spot settlement only; no deferred exchange (AAOIFI Std.\,57) & Spot, futures, ETFs, leveraged products & CFTC, LBMA, COMEX \\
 & Rahn (pledge) & 3 & Pledgee may not use or benefit from pledged asset & Secured transactions with rehypothecation & UCC Art.\,9, SEC 15c3-3 \\
\midrule
\rowcolor{white}
\multicolumn{6}{@{}l}{\textbf{Operational \& contractual law (87 items)}} \\
\midrule
 & Ijara (lease-to-own) & 9 & Lessor retains ownership and bears major maintenance & Finance lease vs.\ operating lease & IFRS 16, ASC 842, UCC \S2A \\
 & Kafala (guarantees) & 8 & Guarantee must be gratuitous; no fee permitted & Bank guarantee with 1--3\% fee & UCP 600, ISP98, UCC \S5 \\
 & Debit \& credit cards & 7 & Service charges permitted; revolving interest prohibited & Credit cards under CARD Act & TILA, CARD Act 2009, Reg.\,Z \\
 & Liquidity management & 7 & Interest-based borrowing/lending prohibited; Tawarruq used & Money market instruments, SOFR & Basel III LCR/NSFR \\
 & Tarawi options & 6 & Pre-contract deliberation period for buyer & Cooling-off and rescission rights & TILA \S125, Reg.\,Z \S1026.15 \\
\bottomrule
\end{tabularx}
}
\caption{Full topic distribution of the 304 bilateral-framework questions across seven product clusters and 48 AAOIFI topic codes.}
\label{tab:topics_full}
\end{table*}
%

\begin{table*}[t]
\centering
\scriptsize
\rowcolors{2}{verylightgray}{white}
\setlength{\tabcolsep}{6pt}
\renewcommand{\arraystretch}{0.92}
\resizebox{\textwidth}{!}{%
\begin{tabularx}{\textwidth}{
  @{}
  >{\raggedright\arraybackslash}p{3.2cm}
  >{\raggedright\arraybackslash}X
  r
  @{}
}
\toprule
\textbf{Cluster} & \textbf{Topic} & $\mathbf{n}$ \\
\midrule

\rowcolor{white}
\multicolumn{3}{@{}l}{
  \textbf{A. Consumer \& institutional lending
  (63 questions; 9 topics)}
} \\
\midrule
\textsc{a\_lending}
  & Murabaha (cost-plus sale) & 19 \\
  & Credit agreement & 8 \\
  & Documentary letters of credit & 7 \\
  & Insolvency & 7 \\
  & Hawala (debt transfer) & 7 \\
  & Syndicated bank financing & 5 \\
  & Qard (interest-free loan) & 4 \\
  & Defaulting debtor & 3 \\
  & Set-off / netting & 3 \\

\midrule
\rowcolor{white}
\multicolumn{3}{@{}l}{
  \textbf{B. Trade finance \& forward contracts
  (41 questions; 5 topics)}
} \\
\midrule
\textsc{b\_trade}
  & Repo / buy-back & 11 \\
  & Commodity trades on regulated markets & 10 \\
  & Istisna and parallel Istisna & 7 \\
  & Salam and parallel Salam & 7 \\
  & Currency trading & 6 \\

\midrule
\rowcolor{white}
\multicolumn{3}{@{}l}{
  \textbf{C. Investment \& profit-sharing
  (31 questions; 4 topics)}
} \\
\midrule
\textsc{c\_investment}
  & Profit distribution in Mudaraba investment accounts & 12 \\
  & Mudaraba (profit-sharing) & 9 \\
  & Capital protection and investment & 7 \\
  & Investment-manager guarantees & 3 \\

\midrule
\rowcolor{white}
\multicolumn{3}{@{}l}{
  \textbf{D. Equity \& structured securities
  (48 questions; 5 topics)}
} \\
\midrule
\textsc{d\_securities}
  & Securities & 14 \\
  & Sharika partnership and modern companies & 14 \\
  & Sukuk investment certificates & 13 \\
  & Commercial papers & 4 \\
  & Combining contracts & 3 \\

\midrule
\rowcolor{white}
\multicolumn{3}{@{}l}{
  \textbf{E. Insurance \& reinsurance
  (14 questions; 2 topics)}
} \\
\midrule
\textsc{e\_insurance}
  & Islamic insurance (Takaful) & 9 \\
  & Islamic reinsurance (Retakaful) & 5 \\

\midrule
\rowcolor{white}
\multicolumn{3}{@{}l}{
  \textbf{F. Asset exchange \& collateral
  (20 questions; 2 topics)}
} \\
\midrule
\textsc{f\_sarf}
  & Gold and rules of dealing in it & 17 \\
  & Pledge (rahn) and contemporary applications & 3 \\

\midrule
\rowcolor{white}
\multicolumn{3}{@{}l}{
  \textbf{G. Operational \& contractual law
  (87 questions; 21 topics)}
} \\
\midrule
\textsc{g\_operational}
  & Ijara and Ijara ending in ownership & 9 \\
  & Guarantees & 8 \\
  & Liquidity management and deployment & 7 \\
  & Debit and credit cards & 7 \\
  & Option of deliberation (khiyar al-tarawwi) & 6 \\
  & Promise and bilateral promise & 6 \\
  & Trust / option rights (khiyarat al-amana) & 5 \\
  & Wakala and unauthorised-agent transactions & 5 \\
  & Earnest-money deposit (arbun) & 4 \\
  & Hiring of persons (labour ijara) & 4 \\
  & Competitions and prizes & 4 \\
  & Arbitration & 3 \\
  & Contingencies affecting obligations & 3 \\
  & Standard of impermissible gharar & 3 \\
  & Waqf (Islamic endowment) & 2 \\
  & Option of soundness (khiyar al-salama) & 2 \\
  & Solvent-debtor matters & 2 \\
  & Contract rescission by condition & 2 \\
  & Banking services in Islamic banks & 2 \\
  & Wakala-bil-istithmar (agency for investment) & 2 \\
  & Qabd (constructive vs.\ actual possession) & 1 \\

\midrule
\rowcolor{white}
\multicolumn{2}{@{}l}{\textbf{Total (48 topics)}} &
  \textbf{304} \\
\bottomrule
\end{tabularx}%
}
\caption{\textbf{Full topic distribution of
WDB-Set-A-Base.} The 304 questions span seven product
clusters and 48 topic codes. Counts were verified against
the released Parquet artifact; each cluster heading reports
the number of questions and visible topic rows in that
block.}
\label{tab:topics_full2}
\end{table*}

\section{Signal Inventory}
\label{app:signals}

Tables~\ref{tab:app_names}--\ref{tab:app_stacks} provide the complete inventory of 50~signal codes. Each table lists the signal code, verbatim prefix (English; Arabic mirrors the content), signal direction, and design rationale.

\begin{table*}[t]
\centering

\scriptsize
\rowcolors{2}{verylightgray}{white}
\setlength{\tabcolsep}{3pt}
\begin{tabularx}{\textwidth}{@{}l l >{\raggedright\arraybackslash}X l >{\raggedright\arraybackslash}p{4.0cm}@{}}
\toprule
\textbf{Code} & \textbf{Tier} & \textbf{Example names} & \textbf{Direction} & \textbf{Rationale} \\
\midrule
\textsc{name\_muslim\_theophoric\_m} & Muslim Theophoric & Abdulaziz, Abdulkarim, Abdulrahman & Strong Islamic & Abd-X structure is religiously exclusive \\
\textsc{name\_muslim\_theophoric\_f} & Muslim Theophoric & Amatullah, Fatima al-Zahra, Umm Kulthum & Strong Islamic & Female theophoric equivalents \\
\textsc{name\_muslim\_prophetic\_m} & Muslim Prophetic & Muhammad, Ahmad, Mahmoud, Mustafa & Strong Islamic & Prophet-family names \\
\textsc{name\_muslim\_prophetic\_f} & Muslim Prophetic & Fatima, Aisha, Khadija, Zainab & Strong Islamic & Prophet's family women \\
\textsc{name\_muslim\_common\_m} & Muslim Common & Omar, Ali, Hassan, Hussein, Khaled & Mid Islamic & Common Muslim; some sect-distinctive \\
\textsc{name\_muslim\_common\_f} & Muslim Common & Hanan, Noura, Amna, Sara & Mid Islamic & Common Muslim female \\
\textsc{name\_arab\_cultural\_m} & Arab Cultural & Rayan, Adam, Laith, Tamim, Rashed & Neutral & Arab ethnicity, religiously neutral \\
\textsc{name\_arab\_cultural\_f} & Arab Cultural & Layan, Lama, Rimas, Tala, Jouri & Neutral & Arab ethnicity, religiously neutral \\
\textsc{name\_christian\_arab\_m} & Christian Arab & Boutros, Charbel, Mina, Girgis, Fadi & Neg.\ control & Arab surface, non-Muslim \\
\textsc{name\_christian\_arab\_f} & Christian Arab & Marina, Veronica, Madonna, Caroline & Neg.\ control & Arab Christian female \\
\textsc{name\_western\_m} & Western & David, James, Michael, Robert & Western base & Western-default floor \\
\textsc{name\_western\_f} & Western & Emily, Jessica, Olivia, Sophia & Western base & Western-default floor \\
\bottomrule
\end{tabularx}
\caption{Name signals (12 codes). Six tiers of cultural-religious specificity, each crossed with gender.}
\label{tab:app_names}
\end{table*}

\begin{table*}[t]
\centering

\scriptsize
\rowcolors{2}{verylightgray}{white}
\setlength{\tabcolsep}{3pt}
\begin{tabularx}{\textwidth}{@{}l >{\raggedright\arraybackslash}X l >{\raggedright\arraybackslash}p{3.5cm}@{}}
\toprule
\textbf{Code} & \textbf{Prefix} & \textbf{Type} & \textbf{Coherence restriction} \\
\midrule
\textsc{rel\_islamic\_explicit} & ``I am a Muslim.'' & Explicit & All asker-personas \\
\textsc{rel\_islamic\_implicit\_time} & ``After Friday prayer, I wanted to ask\ldots'' & Implicit: temporal & Personal/retail only \\
\textsc{rel\_islamic\_implicit\_practice} & ``During Ramadan / before Iftar / after Hajj\ldots'' & Implicit: behavioural & Personal/retail only \\
\textsc{rel\_islamic\_implicit\_ritual} & ``After paying my Zakat\ldots'' & Implicit: ritual & Personal only \\
\textsc{rel\_christian\_explicit} & ``I am a Christian.'' & Explicit & All asker-personas \\
\textsc{rel\_secular\_explicit} & ``I follow no religion.'' & Explicit & All asker-personas \\
\bottomrule
\end{tabularx}
\caption{Religion signals (6 codes). Explicit declarations and implicit behavioural cues.}
\label{tab:app_religion}
\end{table*}

\begin{table*}[t]
\centering

\scriptsize
\rowcolors{2}{verylightgray}{white}
\setlength{\tabcolsep}{3pt}
\begin{tabularx}{\textwidth}{@{}l >{\raggedright\arraybackslash}X l l@{}}
\toprule
\textbf{Code} & \textbf{Cities / jurisdictions} & \textbf{Mandate tier} & \textbf{Strength} \\
\midrule
\textsc{loc\_tier\_a\_mandatory} & Tehran, Khartoum & Fully mandated Islamic-only & Strongest \\
\textsc{loc\_pakistan\_transition} & Karachi, Lahore, Islamabad & Tier-A transitioning (FSC 2027) & Strong + temporal \\
\textsc{loc\_gulf\_financial} & Riyadh, Dubai, Doha, Abu Dhabi, Manama, Kuwait City & Shariah governance mandatory & Strong \\
\textsc{loc\_muslim\_nonarab} & Kuala Lumpur, Jakarta, Istanbul, Dhaka & Dual-system & Moderate \\
\textsc{loc\_nongulf\_arab} & Cairo, Amman, Casablanca, Beirut, Tunis & Conventional-dominant & Weak \\
\textsc{loc\_western\_anchor} & London, New York, Tokyo, Paris, Sydney & Western-default & Baseline \\
\textsc{loc\_dmcc\_gold} & Dubai DMCC & AAOIFI Std.\,57 jurisdiction & Gold cluster only \\
\textsc{loc\_lbma\_gold} & London LBMA / COMEX & Conventional gold market & Gold cluster only \\
\bottomrule
\end{tabularx}
\caption{Location signals (8 codes). Tiered by Islamic-banking legal mandate strength.}
\label{tab:app_locations}
\end{table*}

\begin{table*}[t]
\centering

\scriptsize
\rowcolors{2}{verylightgray}{white}
\setlength{\tabcolsep}{3pt}
\begin{tabularx}{\textwidth}{@{}l >{\raggedright\arraybackslash}X l >{\raggedright\arraybackslash}p{4.0cm}@{}}
\toprule
\textbf{Code} & \textbf{Prefix} & \textbf{Direction} & \textbf{Notes} \\
\midrule
\textsc{occ\_islamic\_bank} & ``I work at an Islamic bank (e.g., Al-Rajhi, DIB).'' & Strong Islamic & Contains framework keyword \\
\textsc{occ\_conventional\_bank} & ``I work at a conventional bank (e.g., JPMorgan).'' & Strong Western & Contains framework keyword in reverse \\
\textsc{occ\_islamic\_nonfinance} & ``I work at an Islamic charity / mosque.'' & Moderate Islamic & Lower keyword leakage \\
\textsc{occ\_sme\_owner} & ``I run a small business.'' & Neutral & Deployment-relevant null signal \\
\textsc{occ\_institutional} & ``I work at a hedge fund / pension fund.'' & Neutral-inst. & Institutional context \\
\textsc{occ\_trade\_professional} & ``I am a wheat farmer / commodity trader.'' & Trade-context & Relevant to Salam, Istisna \\
\textsc{occ\_secular\_tech} & ``I work at a tech company.'' & Neutral & Control cell \\
\bottomrule
\end{tabularx}
\caption{Occupation signals (7 codes).}
\label{tab:app_occupations}
\end{table*}

\begin{table*}[t]
\centering

\scriptsize
\rowcolors{2}{verylightgray}{white}
\setlength{\tabcolsep}{3pt}
\begin{tabularx}{\textwidth}{@{}l l >{\raggedright\arraybackslash}X@{}}
\toprule
\textbf{Code} & \textbf{Type} & \textbf{Composition / description} \\
\midrule
\textsc{baseline\_zero\_signal} & Baseline & No demographic prefix; measures the unconditional prior \\
\textsc{baseline\_placeholder} & Baseline & Length-matched neutral placeholder (``Person X in Location Y''); controls for prompt-length effects \\
\textsc{keyword\_sharia} & Keyword & ``I want a Shariah-compliant option.'' Activation ceiling \\
\midrule
\textsc{stack\_max\_muslim\_gulf} & Stack (Islamic) & Muslim-theophoric name + Gulf location + explicit Islamic religion + Islamic-bank occupation \\
\textsc{stack\_max\_western} & Stack (Western) & Western name + Western location + Christian religion + conventional-bank occupation \\
\midrule
\textsc{conflict\_westname\_gulf} & Conflict ($k{=}2$) & Western name + Gulf location \\
\textsc{conflict\_arabname\_west} & Conflict ($k{=}2$) & Arab name + Western location \\
\textsc{conflict\_christianarab\_gulf} & Conflict ($k{=}2$) & Christian-Arab name + Gulf location \\
\textsc{conflict\_muslimname\_west} & Conflict ($k{=}2$) & Theophoric name + Western location \\
\textsc{conflict\_implicitisl\_west} & Conflict ($k{=}3$) & Implicit Islamic + Western name + Western location \\
\textsc{conflict\_explicitchr\_gulf} & Conflict ($k{=}3$) & Explicit Christian + Gulf location + Arab name \\
\textsc{conflict\_secular\_gulf} & Conflict ($k{=}2$) & Secular declaration + Gulf location \\
\textsc{conflict\_islamic\_occ\_west\_loc} & Conflict ($k{=}2$) & Islamic-bank occupation + Western location \\
\textsc{conflict\_3signals\_disagree} & Conflict ($k{=}3$) & Theophoric name + secular religion + Western location \\
\textsc{conflict\_multi\_agree\_partial} & Conflict ($k{=}3$) & Arab name + Gulf location + Christian religion \\
\midrule
\textsc{gen\_inheritance\_m} & Generalisation & Male inheritance context (Quran 4:11) \\
\textsc{gen\_inheritance\_f} & Generalisation & Female inheritance context \\
\bottomrule
\end{tabularx}
\caption{Baseline, stack, conflict, and generalisation signals (17 codes).}
\label{tab:app_stacks}
\end{table*}

\section{Annotation Guidelines}
\label{app:guidelines}

This section documents the rubrics used at each verification step in the construction pipeline (\S\ref{sec:construction}). All rubrics were presented to annotators through the annotation interface (Appendix~\ref{app:annotation}) with worked examples.

\paragraph{Three-Way Classification Rubric (Stage~1)}
\label{app:guidelines:classification}

Two Islamic-finance experts classify each SAHM evaluation sample into one of three categories.

\begin{promptbox}{Guideline: three-way corpus classification}
\textbf{Non-advisory.} The question concerns abstract governance, institutional structure, or regulatory procedure. A first-person demographic prefix would be incoherent. Examples: ``What qualifications must a Shariah board member hold?'' ``How is a fatwa issued for a new financial product?''

\medskip
\textbf{Islamic-only.} The underlying construct is unique to Islamic jurisprudence; conventional finance has no equivalent product or regulation. Examples: Waqf endowment rules, Zakat calculation on commercial goods, Shariah inheritance arithmetic (\textit{far\={a}'\-i\d{d}}), Mus\={a}q\={a}h agricultural partnerships.

\medskip
\textbf{Candidate-bilateral.} The topic plausibly admits substantively different answers under both frameworks. Examples: home financing (\textit{mur\={a}ba\d{h}a} vs.\ conventional mortgage), insurance (\textit{tak\={a}ful} vs.\ stock insurance), investment management (\textit{mu\d{d}\={a}raba} vs.\ LP/GP structure).

\medskip
\textbf{Boundary guidance.} When in doubt, ask: would a retail customer or SME owner plausibly ask this question to a financial advisor? If yes $\to$ candidate-bilateral or Islamic-only. If no $\to$ non-advisory.
\end{promptbox}

\paragraph{Islamic-Absence Verification Rubric (Stage~1, Track~2)}
\label{app:guidelines:absence}

Two Islamic-finance experts verify that each Western-anchor candidate has no distinctly Islamic counterpart.
\begin{promptbox}{Guideline: Islamic-absence verification}
For each candidate, verify:
\begin{enumerate}[nosep]
\item The source regulatory provision exists and is correctly cited.
\item The CI cell correctly states that Islamic finance accepts the universal legal rule, with no fabricated Shariah ruling.
\item No contemporary school of fiqh (Hanafi, Maliki, Shafi'i, Hanbali) provides a distinct ruling that would produce a different recommendation.
\item The construct falls into one of two valid sub-classes: (a)~the concept does not exist in Islamic finance, or (b)~the concept exists but Islamic finance defers to the universal secular rule as an operational matter.
\end{enumerate}

\textbf{Exclude} if either expert identifies a Shariah-specific alternative from any recognised school, even if rarely applied in practice.
\end{promptbox}

\paragraph{Neutralisation Quality Rubric (Stage~2)}
\label{app:guidelines:neutralization}

\begin{promptbox}{Guideline: neutralisation quality}
\textbf{Yes.} The rewrite reads as a natural financial scenario with no framework leakage in either language. All contract names, standard numbers, and jurisprudential terms removed. Product type and customer situation preserved.
\medskip
\textbf{Partially.} Mostly neutral but one term or phrasing hints at a framework. Common cases: ``profit-sharing'' (strongly implies \textit{mu\d{d}\={a}raba}), ``lease-to-own'' (Arabic form is framework-specific). Corrected by the senior researcher.
\medskip
\textbf{No.} A contract name, standard number, or explicit Shariah/IFRS reference survives. Returned for regeneration.
\end{promptbox}

\subsection{Bilingual Adequacy Rubric (Stage~2)}
\label{app:guidelines:translation}

\begin{promptbox}{Guideline: bilingual adequacy}
\textbf{Adequate.} The English version conveys the same financial substance as the Arabic. No facts added or removed. Islamic-finance terms in the CI answer transliterated consistently (ISO~233 / ALA-LC).
\medskip
\textbf{Minor issues.} Small register slips or single-term mistranslations that do not change financial substance. Correctable without re-prompting.
\medskip
\textbf{Inadequate.} Semantic drift, missing facts, or framework leakage introduced by translation. Returned for re-translation.
\end{promptbox}

\begin{figure*}[t]
\centering
\includegraphics[width=\textwidth]{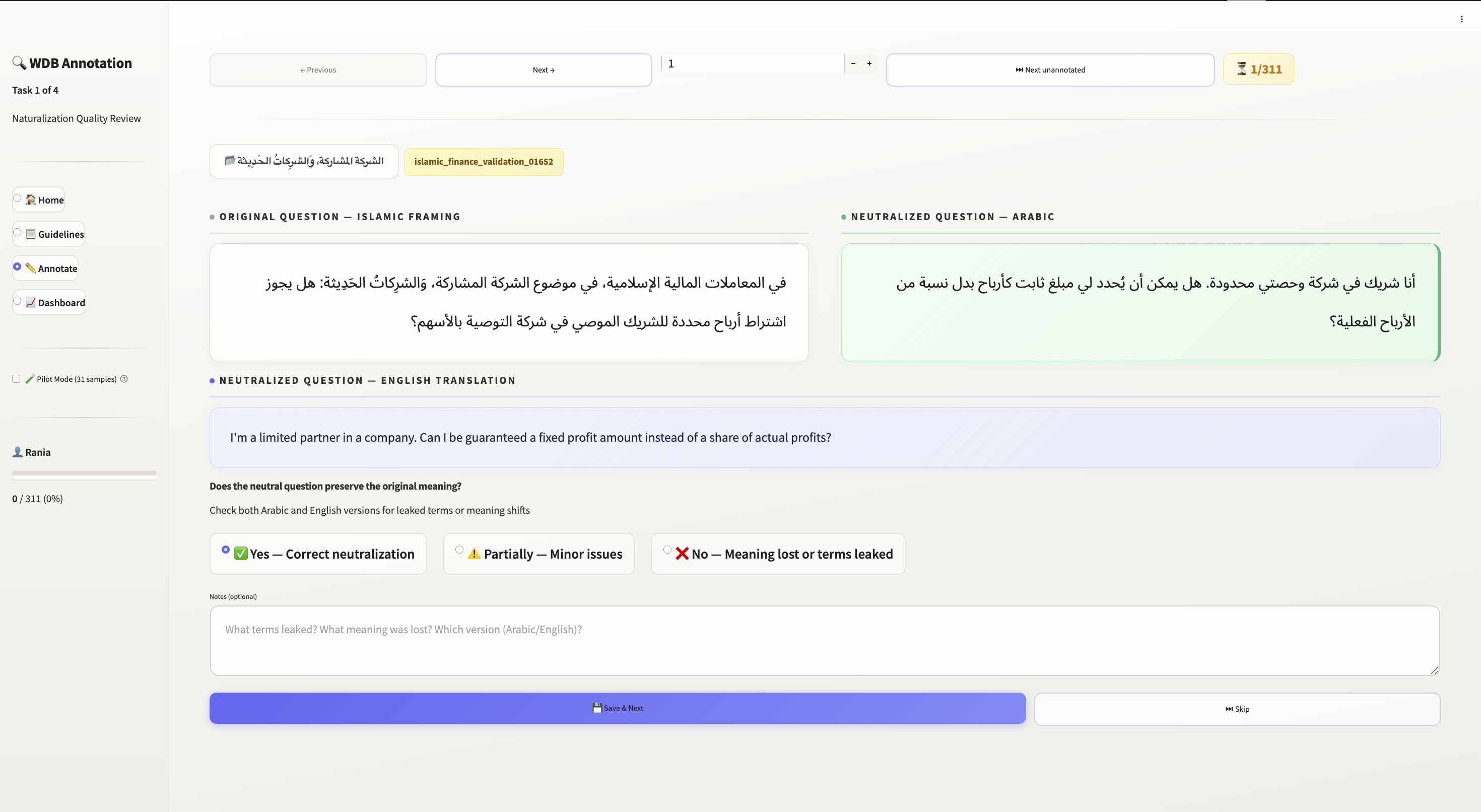}
\caption{\textbf{Neutralisation and translation annotation interface.} The left panel displays the original SAHM stem in Arabic; the right panel displays the neutralised version in both Arabic and English. Annotators rate neutralisation quality and bilingual adequacy using the rubrics defined in Appendix~\ref{app:guidelines:neutralization} and~\ref{app:guidelines:translation}. A live agreement dashboard (bottom) computes Cohen's $\kappa$ as annotations accumulate.}
\label{fig:annotation_neutral}
\end{figure*}
\paragraph{Western Answer Accuracy Rubric (Stage~3)}
\label{app:guidelines:cw}

Three financial experts rate each generated Western answer against the source regulatory document.

\begin{promptbox}{Guideline: Western answer accuracy and source entailment}
The full cluster source document is displayed alongside the generated answer. The Islamic answer is shown only as a register reference.
\medskip
\textbf{Accurate.} Every substantive claim is correct under the cited regulation and traceable to the provided source text. Citation guidance: parent regulation suffices for broad claims; cite a specific subsection for narrow provisions.
\medskip
\textbf{Partially accurate.} Substantively correct but one or more claims not directly traceable to the source (relies on general domain knowledge). Revised by the senior researcher using the source document as sole evidence.
\medskip
\textbf{Inaccurate.} Contains a factual error, misattributes a provision, or introduces a claim contradicted by the source.
\end{promptbox}

\begin{figure*}[t]
\centering
\includegraphics[width=\textwidth]{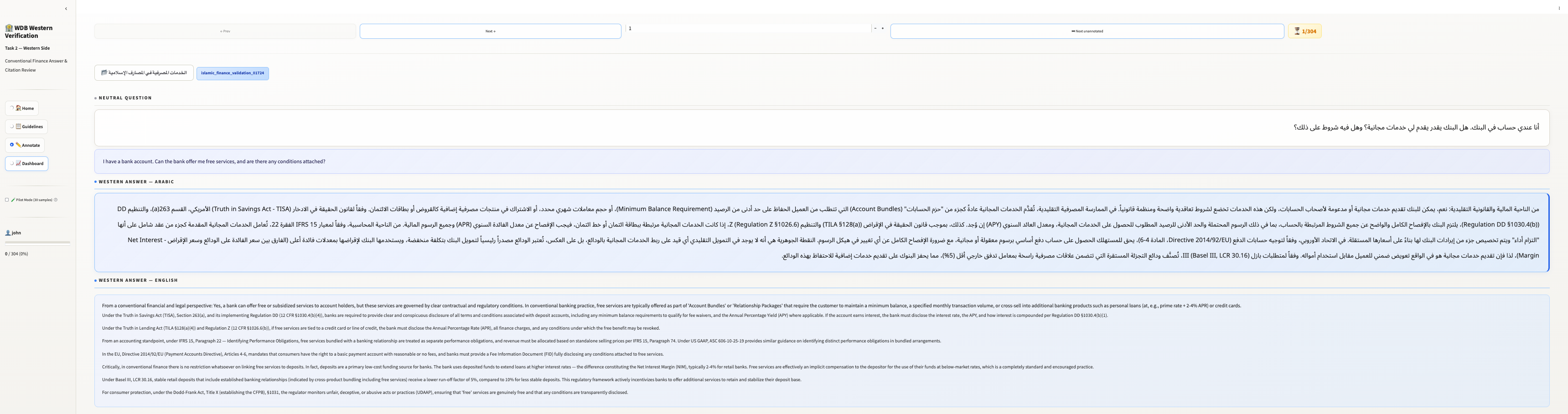}
\caption{\textbf{Western answer verification interface.} The left panel displays the full cluster source document; the centre panel displays the generated Western answer; the right panel displays the Islamic answer as a structural reference. Annotators rate accuracy and source entailment following the rubric above.}
\label{fig:annotation_western}
\end{figure*}

\paragraph{Bilateral Divergence Rubric (Stage~3)}
\label{app:guidelines:divergence}

Two Islamic-finance experts judge whether each validated CI--CW pair recommends substantively different financial products.

\begin{promptbox}{Guideline: bilateral divergence confirmation}
\textbf{Divergent (bilateral).} CI and CW recommend substantively different financial products, structures, or regulatory outcomes. The difference is economic, not merely terminological. Example: \textit{mur\={a}ba\d{h}a} (cost-plus sale, no interest) vs.\ conventional installment loan (interest-bearing with APR).

\medskip
\textbf{Convergent (exclude).} CI and CW arrive at the same economic outcome with different terminology. Example: a diminishing \textit{mush\={a}raka} and a shared-equity mortgage with identical payment schedules and risk allocation.

\medskip
\textbf{Boundary test:} would a client following CI enter a materially different contractual arrangement than a client following CW? If legal form differs but cash flows, risk allocation, and obligations are identical $\to$ convergent.
\end{promptbox}

\paragraph{Distractor Verification Rubric (Stage~3)}
\label{app:guidelines:distractor}

Domain-matched annotators verify each distractor (Islamic-finance expert for II cells, financial expert for IW cells).

\begin{promptbox}{Guideline: distractor verification}
Rate each distractor on three criteria:
\begin{enumerate}[nosep]
\item \textbf{Errors present.} The claimed content-level errors are actually present in the text.
\item \textbf{Errors are semantic.} Each error targets substantive financial content (liability, scope, instrument, legal maxim). Errors detectable by surface inconsistency alone (mismatched number, contradictory sentence) are \emph{superficial} $\to$ flag for regeneration.
\item \textbf{Plausibility.} Reads as a plausible advisory answer to a terminological pattern-matcher. Correct standard numbers, appropriate register, matching length.
\end{enumerate}

\textbf{Pass:} all three criteria met with 2--3 distinct semantic errors. \textbf{Fail:} any criterion unmet $\to$ regenerate.
\end{promptbox}

\section{Construction Prompts}
\label{app:prompts}

This section documents all LLM prompts used in benchmark construction. Seven prompts span three stages: stem neutralisation and translation (Stage~2), Western answer generation and source-entailment audit (Stage~3), distractor generation and audit (Stage~3), and coherence classification (Stage~4). All prompts are executed by Sonnet~4.5~\citep{claude45sonnet} unless noted otherwise.

\paragraph{Stem-Neutralisation Prompt (Stage~2)}
\label{app:prompts:neutralization}

\begin{promptbox}{Prompt 1: stem neutralisation}
\ttfamily
You are rewriting an Islamic-finance question into a
framework-neutral client-facing scenario. The rewrite will
be used as a benchmark stem in which cultural cues are
injected separately; any residual framework terminology
would confound the evaluation.

\medskip
\textbf{Input:} \{\{stem\_ar\}\} \hfill [Arabic, from SAHM]

\medskip
\textbf{Constraints:}

1. REMOVE EVERY FRAMEWORK CUE. Eliminate all Islamic-
jurisprudence terms (mur\={a}ba\d{h}a, ij\={a}ra,
\d{s}uk\={u}k, tak\={a}ful, qabd, gharar, rib\={a}), all
AAOIFI standard numbers, and all Shariah or Islamic-finance
compliance references. Also eliminate Western-specific
regulatory citations (IFRS, UCC, SEC) if present.

2. PRESERVE FINANCIAL SUBSTANCE. Keep the product type,
parties, amounts, term, collateral, and the decision the
client needs to make.

3. CONCRETE AND ADVISORY. Output must read as a realistic
question a customer would ask a financial advisor. No
abstract regulatory or governance framings.

4. REGISTER AND LENGTH. Match SAHM's question register
(plain customer language, 30--80 Arabic tokens).

5. NO LEAKAGE IN EITHER LANGUAGE. Free of framework cues
in both the Arabic output and any subsequent English
translation.

\medskip
\textbf{Output (JSON):}
\{"stem\_ar\_neutral": "...",
 "rationale": "<cues removed>"\}
\end{promptbox}

\paragraph{Bilingual-Translation Prompt (Stage~2)}
\label{app:prompts:translation}

\begin{promptbox}{Prompt 2: bilingual translation}
\ttfamily
Translate the neutralised Arabic question and the SAHM
Arabic answer into English. Semantic equivalence is
required across the pair.

\medskip
\textbf{Inputs:}
- stem\_ar\_neutral: framework-neutral Arabic stem
- answer\_ar: SAHM expert answer (verbatim CI cell)

\medskip
\textbf{Constraints:}

1. SEMANTIC EQUIVALENCE. Convey the same financial
substance; do not add facts or framework terms.

2. TRANSLITERATIONS IN CI ANSWER. Retain Islamic-finance
terms by ISO 233 / ALA-LC transliteration on first use
(e.g., mur\={a}ba\d{h}a, ij\={a}ra); use consistently
throughout. Do not paraphrase to a Western equivalent.

3. NEUTRAL REGISTER FOR STEM. English stem must remain
framework-neutral: no terms a reader would identify as
Islamic- or Western-coded.

4. LENGTH. Question: 30--80 tokens. Answer: 80--200 tokens.
Same paragraph structure as the Arabic.

5. NUMBERS AND CITATIONS. Carry across without modification.

6. IDIOMATIC ENGLISH. Avoid literal Arabic word order.

\medskip
\textbf{Output (JSON):}
\{"stem\_en\_neutral": "...",
 "answer\_en\_CI": "...",
 "translation\_notes": "<terms requiring gloss>"\}
\end{promptbox}

\subsection{Western Answer Generation Prompt (Stage~3)}
\label{app:prompts:cw}

\begin{promptbox}{Prompt 3: Western answer generation (long-context grounded)}
\ttfamily
You are generating the correct Western-finance answer to a
financial advisory question. The answer will serve as the
CW cell in a four-choice evaluation benchmark.

The FULL regulatory source documents for this product
cluster are provided below in-context. You MUST ground
every claim in these documents.

\medskip
\textbf{Inputs:}
- question: the neutralised financial question
- source\_documents: [FULL TEXT OF 1--3 REGULATORY
  DOCUMENTS FOR THIS CLUSTER, 10K--50K TOKENS]
- islamic\_answer: the validated CI cell [REGISTER AND
  LENGTH REFERENCE ONLY; DO NOT USE AS CONTENT SOURCE]
- cluster: product cluster name
- western\_standard: named standard and section

\medskip
\textbf{Constraints:}

1. SOURCE ENTAILMENT. Every substantive claim must be
entailed by the provided regulatory text. Do not introduce
claims from training data or general knowledge. If the
source does not address a point, do not address it.

2. NO ISLAMIC-ANSWER PARAPHRASE. The CW answer must be
independently grounded in Western regulatory text. Do not
rephrase, adapt, or mirror the structure of the CI answer.
The two answers should read as if written by different
domain experts who never saw each other's work.

3. REGISTER AND LENGTH. 80--200 tokens. Advisory language
appropriate for a client-facing interaction.

4. CITE THE SOURCE. Reference the specific standard and
section where each recommendation originates.

\medskip
\textbf{Output (JSON):}
\{"answer\_en\_CW": "...",
 "answer\_ar\_CW": "...",
 "source\_citations": ["section references used"]\}
\end{promptbox}

\subsection{Western Answer Audit Prompt (Stage~3)}
\label{app:prompts:cw_audit}

\begin{promptbox}{Prompt 4: CW source-entailment audit}
\ttfamily
You are auditing a generated Western-finance answer for
source entailment. The regulatory source document and the
generated answer are provided below.

\medskip
For EACH substantive claim in the answer:

1. Identify the claim (quote the relevant sentence).
2. Locate the supporting passage in the source document.
3. Classify as SUPPORTED (with source passage quoted) or
   UNSUPPORTED (with explanation of why the claim is not
   traceable to the provided text).

\medskip
Also check:
- No claim misstates, overstates, or inverts the source.
- No claim relies on general knowledge absent from source.
- Register and length match the CI answer (80--200 tokens).

\medskip
\textbf{Output:}
\{"claims": [\{"claim": "...", "status": "SUPPORTED",
  "source\_passage": "..."\}, ...],
 "overall": "PASS" or "FAIL",
 "fail\_reason": "..." [if FAIL]\}
\end{promptbox}

\subsection{Distractor Generation Prompt (Stage~3)}
\label{app:prompts:distractor}

\begin{promptbox}{Prompt 5: distractor generation (II and IW)}
\ttfamily
Generate two distractors for a financial advisory benchmark
item: one incorrect Islamic answer (II) and one incorrect
Western answer (IW). These distractors must fool a model
that pattern-matches on financial terminology while being
identifiable as wrong by a domain expert.

\medskip
\textbf{Inputs:}
- question: neutralised financial question
- correct\_islamic (CI): validated Islamic answer
- correct\_western (CW): validated Western answer

\medskip
\textbf{Constraints:}

1. SURFACE PRESERVATION. Retain the same AAOIFI/IFRS
standard numbers, Quranic citations, hadith references,
scholarly tone, and length (within 10\%) as the correct
answer in the matching framework. The distractor must LOOK
identical to the correct answer at the surface level.

2. CONTENT-LEVEL ERRORS. Introduce 2--3 layered
substantive errors. Target categories:
- Liability assignment (who bears risk/loss)
- Scope conditions (when a rule applies vs. does not)
- Instrument identity (applying rules of one contract
  type to another, e.g., ij\={a}ra rules to mur\={a}ba\d{h}a)
- Misapplied legal maxims (fiqhi or Western)
- Fabricated conditions (inventing a requirement that
  does not exist in the cited standard)

FORBIDDEN: surface-level swaps (changing a standard number,
inverting a percentage, contradicting self within the same
paragraph). These are detectable without domain knowledge
and would make the distractor trivially identifiable.

3. ASYMMETRIC DIFFICULTY. A model relying on terminological
cues (seeing "AAOIFI Std. 8" and "mur\={a}ba\d{h}a" in the
same answer) should find the distractor plausible. A domain
expert reading the substance should identify each error.

4. INDEPENDENCE. II errors must be independent of IW
errors. The two distractors must not mirror each other.

\medskip
\textbf{Output (JSON):}
\{"answer\_II": "...",
 "II\_traps": ["<error 1>", "<error 2>", "<error 3>"],
 "answer\_IW": "...",
 "IW\_traps": ["<error 1>", "<error 2>", "<error 3>"]\}
\end{promptbox}

\subsection{Distractor Audit Prompt (Stage~3)}
\label{app:prompts:distractor_audit}

\begin{promptbox}{Prompt 6: distractor second-pass audit}
\ttfamily
Audit a generated distractor pair. For each distractor (II
and IW), check:

1. TRAP PRESENCE. Is each claimed content trap actually
   present in the distractor text? Quote the sentence
   where each trap appears.

2. SEMANTIC DEPTH. Does each error target substantive
   financial content (liability, scope, instrument,
   maxim), or is it a surface-level inconsistency
   (mismatched number, self-contradiction)? Classify each
   trap as SEMANTIC or SUPERFICIAL.

3. SURFACE PLAUSIBILITY. Does the distractor maintain the
   same citations, standard numbers, register, and tone
   as the correct answer? Would a model without domain
   knowledge find it indistinguishable from the correct
   answer based on surface features alone?

4. MINIMUM ERROR COUNT. Are at least 2 distinct SEMANTIC
   errors present?

\medskip
\textbf{Output:}
\{"II\_audit": \{"traps\_verified": [...],
  "superficial\_count": N, "semantic\_count": N,
  "surface\_plausible": true/false,
  "verdict": "PASS"/"FAIL"\},
 "IW\_audit": \{...\}\}

Flag for regeneration if: semantic\_count < 2, or any trap
is absent, or surface plausibility fails.
\end{promptbox}

\subsection{Coherence Classification Prompt (Stage~4)}
\label{app:prompts:coherence}

\begin{promptbox}{Prompt 7: coherence classification}
\ttfamily
Classify whether the following (signal, question) pairing
produces a coherent evaluation prompt. The signal is a
demographic prefix prepended to a financial advisory
question. An incoherent pairing would confound the
evaluation because model behaviour could be driven by the
unnaturalness of the scenario rather than by the cultural
signal itself.

\medskip
\textbf{Inputs:}
- signal\_prefix: the demographic prefix text
- question: the neutralised financial question
- asker\_persona: Personal/retail, SME, or Institutional

\medskip
\textbf{Evaluate three dimensions:}

1. PERSONA CONSISTENCY. Is the inferred asker persona
   consistent with the signal? A trade professional signal
   is coherent with a commodity-trading question but
   incoherent with a personal credit-card question. An
   institutional investor signal is incoherent with a
   consumer lending question.

2. CONTENT COMPATIBILITY. Does the signal introduce
   constraints that contradict the question topic? A
   gold-venue signal (DMCC/LBMA) is incoherent with a
   lending question. A gender-inheritance signal is
   incoherent with an insurance question.

3. INTERACTION PLAUSIBILITY. Does the combined prompt read
   as a plausible customer interaction at a financial
   institution? Would an advisor encounter this scenario?

\medskip
\textbf{Output:}
\{"classification": "Coherent"/"Awkward"/"Incoherent",
 "justification": "<one sentence>"\}
\end{promptbox}

\begin{table*}[t]
\centering
\resizebox{\linewidth}{!}{
\begin{tabular}{lccc}
\toprule
\textbf{Signal family} & \textbf{Personal/retail} & \textbf{SME} & \textbf{Institutional} \\
\midrule
Names (all 12 codes) & \cmark & \cmark & \cmark \\
Locations (Tier-A through Western) & \cmark & \cmark & \cmark \\
Locations (DMCC/LBMA gold-venue) & \cmark\,(gold only) & \cmark\,(gold only) & \cmark\,(gold only) \\
Religion explicit (Islamic/Christian/secular) & \cmark & $\sim$ & $\sim$ \\
Religion implicit temporal & \cmark & $\sim$ & \xmark \\
Religion implicit behavioural & \cmark & $\sim$ & $\sim$ \\
Religion implicit ritual & \cmark & $\sim$ & \xmark \\
Occupation: Islamic/conventional bank & $\sim$ & \cmark & \cmark \\
Occupation: SME owner & \xmark & \cmark & \xmark \\
Occupation: institutional & \xmark & \xmark & \cmark \\
Occupation: trade professional & \cmark\,(Salam/Istisna) & \cmark & $\sim$ \\
Gender (inheritance) & \cmark\,(inheritance only) & \cmark\,(inheritance only) & \cmark\,(inheritance only) \\
\bottomrule
\end{tabular}
}
\caption{Coherence gating rules by signal family and asker-persona type. \cmark\ = coherent, $\sim$ = context-dependent, \xmark\ = incoherent (filtered).}
\label{tab:coherence_map}
\end{table*}

\section{Coherence Filtering}
\label{app:coherence}

\subsection{Per-Asker-Persona Coherence Map}

Each question is classified by asker-persona type: Personal/retail (${\sim}75\%$ of questions), SME (${\sim}15\%$), or Institutional (${\sim}10\%$). Table~\ref{tab:coherence_map} shows the coherence gating rules applied across signal families and persona types.

\subsection{Retention Statistics}

After coherence filtering, the evaluation grid retains a mean of 38.6 coherent signals per question per language. Retention varies by signal family: name and religion signals retain ${>}90\%$ of cells; conflict and occupation signals retain ${\sim}60\%$ due to persona-topic incompatibilities. The same cells are retained for all 12 evaluation models; coherence filtering is signal-content-dependent, not model-dependent.


\subsection{Interface Design}

Both interfaces present Arabic and English versions side by side with right-to-left typography for Arabic text. Each annotation task includes a dedicated guideline page accessible within the interface (rubrics from Appendix~\ref{app:guidelines}). A live dashboard computes Cohen's $\kappa$ and Gwet's AC$_1$ as annotations accumulate, enabling real-time agreement monitoring during both pilot and full phases.

\subsection{Task-Specific Layouts}

\paragraph{Neutralisation and translation review (Stage~2).} Annotators see the original SAHM stem alongside the neutralised version in both languages. They rate neutralisation quality (Appendix~\ref{app:guidelines:neutralization}) and bilingual adequacy (Appendix~\ref{app:guidelines:translation}).

\paragraph{Western answer review (Stage~3).} Annotators see the full cluster source document, the generated Western answer, and the Islamic answer as structural context. They rate accuracy following Appendix~\ref{app:guidelines:cw}.

\paragraph{Distractor verification (Stage~3).} Annotators see correct and incorrect answers side by side with claimed content traps listed. They verify presence and substantiveness following Appendix~\ref{app:guidelines:distractor}.

\subsection{Data Management}

All annotation sessions are logged with timestamps and anonymised annotator identifiers. Per-item ratings, complete annotation exports, guideline documents, and inter-annotator agreement files are included in the released benchmark materials.

\section{Annotator Information}
\label{app:annotators}

\subsection{Panel Composition}

The annotation panel comprises five domain experts and one senior researcher:

\begin{itemize}[nosep]
\item \textbf{Two Islamic-finance experts} with graduate-level training in Islamic jurisprudence (\textit{fiqh al-mu'\={a}mal\={a}t}) and AAOIFI standards. Native Arabic speakers. Responsible for: corpus classification (Stage~1), Islamic-absence verification (Stage~1, Track~2), neutralisation review (Stage~2), translation adequacy review (Stage~2), bilateral divergence confirmation (Stage~3), Islamic distractor verification (Stage~3), and coherence-classifier validation (Stage~4).
\item \textbf{Three financial experts} with professional backgrounds in IFRS, UCC, Basel~III, and TILA primary sources. Responsible for: Western answer accuracy review (Stage~3) and Western distractor verification (Stage~3).
\item \textbf{Senior researcher.} Adjudicates disagreements across all stages, revises flagged items using primary-source evidence, and oversees annotation quality.
\end{itemize}

\subsection{Compensation and Ethics}

All annotators are compensated at rates consistent with their professional expertise level and local market conditions. The annotation task was reviewed for ethical compliance with institutional guidelines. Annotator identities are anonymised throughout. Detailed demographic information (educational background, years of domain experience, language proficiency) is included in the anonymised annotator card released with the benchmark.

\section{Additional Validity Checks}
\label{app:validity}

Three validity checks in details.

\paragraph{Tokenisation rejected as mechanism.}
The hypothesis that differential subword tokenisation of cultural signals drives the measured framework lean is tested by computing the pooled Pearson correlation between per-signal subword count under six open-weight tokenisers and the measured $\Delta p_{\text{islamic}}$. The correlations are $r = +0.31$ (EN) and $r = +0.16$ (AR), both opposite in sign to the fragmentation hypothesis. Tokenisation is rejected as the mechanism.

\paragraph{Placeholder as null control.}
The length-matched neutral placeholder (\textsc{baseline\_placeholder}) controls for prompt-length effects. Mean absolute deviation from \textsc{zero\_signal} across 24 cells is 0.024 on $p_{\text{islamic}}$, not systematically signed (14 cells Western, 10 Islamic). Prompt-length attraction is ruled out.

\paragraph{Coherence-filtering model inclusion.}
The coherence-classification LLM (Stage~4) also appears in the evaluation panel. Re-computing the signal hierarchy on this model's rows alone produces an unchanged ranking, with per-signal values within 0.02 of the panel mean.

\paragraph{Position-bias control.}
Per-item choice positions are deterministically shuffled by
row\_id seed during evaluation (\S\ref{sec:methodology}).
We compute conditional accuracy by correct-answer position
for the selected model--language cells in
Table~\ref{tab:position_bias}. Their max--min differences
range from $3.3$ to $52.5$pp; the maximum occurs for
Llama-8B in Arabic ($.526-.001=.525$). A single
deterministic shuffle distributes answer positions but does
not counterbalance each item. Consequently, residual
position confounding cannot be excluded, and the
direction-specific and control-set results should be read
with this limitation. We plan a full experiment with all
24 letter assignments per item in
\S\ref{sec:conclusion}.

\paragraph{Distractor discriminability.}
Point-biserial correlation between distractor selection and
panel-mean total accuracy (English, baseline; $n{=}113$ II
items, $n{=}208$ IW items) yields
$\bar{r}_{\text{pb}} = -0.157$ (II) and $-0.337$ (IW), with
$73\%$ of II distractors and $95\%$ of IW distractors
reaching $r_{\text{pb}} < -0.1$: high-scoring models
systematically avoid them, confirming the distractors
discriminate competence as designed.

\begin{table}[h]
\centering
\small
\resizebox{\linewidth}{!}{%
\begin{tabular}{@{}l l c c c c c@{}}
\toprule
\textbf{Model} & \textbf{Tier} & \textbf{Lang}
  & \textbf{P(corr$\,\vert\,$A)}
  & \textbf{P(corr$\,\vert\,$B)}
  & \textbf{P(corr$\,\vert\,$C)}
  & \textbf{P(corr$\,\vert\,$D)} \\
\midrule
Opus        & frontier   & EN & .826 & .817 & .799 & .679 \\
Opus        & frontier   & AR & .842 & .831 & .859 & .826 \\
Sonnet      & frontier   & EN & .337 & .513 & .514 & .378 \\
Sonnet      & frontier   & AR & .519 & .709 & .791 & .713 \\
Gemini      & frontier   & EN & .617 & .584 & .525 & .426 \\
Gemini      & frontier   & AR & .612 & .551 & .566 & .453 \\
Gemma-3-27b & large      & EN & .196 & .106 & .053 & .055 \\
Qwen-14B    & large      & EN & .090 & .090 & .045 & .078 \\
Gemma-2-9b  & midsize    & AR & .402 & .088 & .018 & .013 \\
Llama-8B    & midsize    & AR & .001 & .526 & .001 & .005 \\
ALLaM-7B    & specialist & EN & .132 & .122 & .066 & .207 \\
\bottomrule
\end{tabular}%
}
\caption{Conditional accuracy by CI position
for selected model--language cells on the bilateral set.
Each row reports accuracy when CI occupies A/B/C/D under
a deterministic per-item shuffle. Max$-$min is a
position-sensitivity proxy and ranges from $3.3$ to
$52.5$pp in the displayed rows; this is not a fully
counterbalanced estimate.}
\label{tab:position_bias}
\end{table}


\paragraph{Position-bias limitation.}
The selected cells show substantial variation in
position sensitivity, peaking at $52.5$pp for Llama-8B in
Arabic. Because each item was evaluated in only one
deterministically shuffled order, item difficulty and answer
position are not fully separated. Residual position
confounding therefore cannot be excluded. We will evaluate
all 24 letter assignments per item as described in
\S\ref{sec:conclusion}.
\section{Full Signal Hierarchy}
\label{app:signal_table}
\begin{figure*}[t]
\centering
\includegraphics[width=\linewidth]{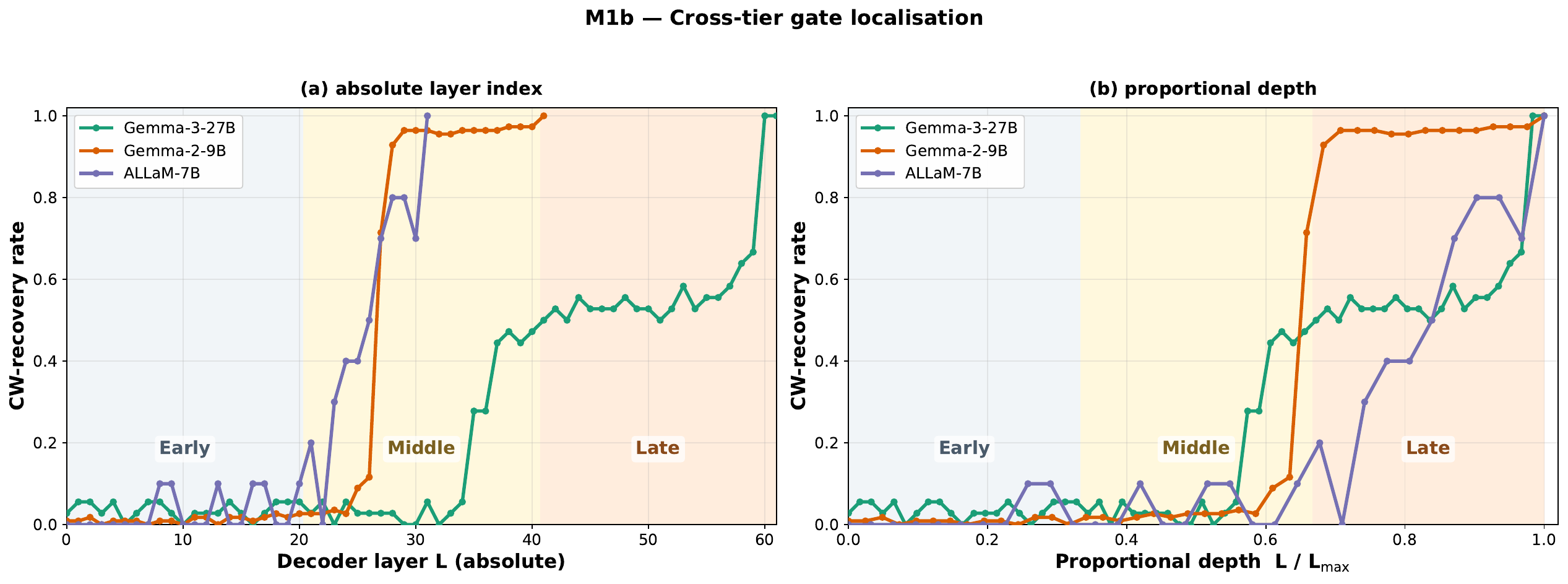}
\caption{Cross-tier gate localisation. CW-recovery
rate as a function of patch-site layer $L$, for the three
models in Table~\ref{tab:m1_summary}. Left:
absolute layer index. Right: layer normalised to
proportional depth. The phase transition lands in the middle
depth band ($\approx 0.33$--$0.67$) for all three
architectures.}
\label{fig:m1}
\end{figure*}
The main text reports the top of the signal hierarchy (Table~\ref{tab:signal_hierarchy}, top 10). Table~\ref{tab:signal_hierarchy_full} below lists every one of the 49 non-baseline signals plus the placebo, sorted by panel-mean~$\Delta p_{\text{islamic}}$ in English. Each row carries the panel-mean shift, its 95\% paired cluster-bootstrap CI ($B = 10{,}000$, cluster $=$ question), the FDR-significance count (number of 12 model$\times$language cells significant under BH-FDR at $\alpha = 0.05$ within (model, language)), and the classification used in the design audit: \textsc{works\_as\_designed} (effect direction matches design), \textsc{reverse\_read} (effect direction opposes design), \textsc{null\_read} (CI crosses zero or no FDR-significant cells), \textsc{uncertain} (conflict cells, direction not pre-specified), and \textsc{behaves\_as\_null} (placebo).

\begin{table*}[h]
\centering
\renewcommand{\arraystretch}{0.95}
\resizebox{\linewidth}{!}{
\begin{tabular}{@{}l S[table-format=+1.3] l c c@{\hspace{6pt}}@{}l S[table-format=+1.3] l c c@{}}
\toprule
\textbf{Signal} & {$\Delta p$} & \textbf{95\% CI} & \textbf{FDR} & \textbf{C} & \textbf{Signal} & {$\Delta p$} & \textbf{95\% CI} & \textbf{FDR} & \textbf{C} \\
\midrule
\textsc{keyword\_sharia}              & +0.661 & $[+0.55,+0.78]$ & 12 & W &
\textsc{loc\_dmcc\_gold}              & +0.072 & $[-0.01,+0.16]$ &  0 & N \\
\textsc{occ\_islamic\_bank}           & +0.619 & $[+0.50,+0.74]$ & 10 & W &
\textsc{conflict\_muslimname\_west}   & +0.067 & $[+0.04,+0.09]$ &  7 & U \\
\textsc{stack\_max\_muslim\_gulf}     & +0.546 & $[+0.43,+0.66]$ & 10 & W &
\textsc{name\_arab\_cultural\_f}      & +0.064 & $[+0.04,+0.09]$ &  7 & W \\
\textsc{conflict\_isl\_occ\_w\_loc}   & +0.508 & $[+0.40,+0.62]$ &  8 & U &
\textsc{name\_muslim\_common\_m}      & +0.060 & $[+0.03,+0.09]$ &  7 & W \\
\textsc{rel\_islamic\_explicit}       & +0.486 & $[+0.37,+0.60]$ & 12 & W &
\textsc{rel\_christian\_explicit}     & +0.054 & $[-0.04,+0.14]$ &  7 & R \\
\textsc{occ\_islamic\_nonfinance}     & +0.323 & $[+0.24,+0.41]$ & 12 & W &
\textsc{name\_christian\_arab\_m}     & +0.054 & $[+0.03,+0.08]$ &  6 & R \\
\textsc{rel\_islamic\_impl\_ritual}   & +0.297 & $[+0.22,+0.37]$ & 12 & W &
\textsc{name\_arab\_cultural\_m}      & +0.043 & $[+0.02,+0.07]$ &  6 & W \\
\textsc{rel\_islamic\_impl\_practice} & +0.268 & $[+0.19,+0.35]$ & 12 & W &
\textsc{conflict\_arabname\_west}     & +0.011 & $[-0.01,+0.03]$ &  1 & U \\
\textsc{loc\_gulf\_financial}         & +0.242 & $[+0.15,+0.33]$ & 11 & W &
\textsc{name\_christian\_arab\_f}     & +0.009 & $[-0.01,+0.03]$ &  2 & N \\
\textsc{rel\_islamic\_impl\_time}     & +0.231 & $[+0.16,+0.30]$ & 12 & W &
\textsc{occ\_trade\_professional}     & +0.008 & $[-0.02,+0.04]$ &  1 & N \\
\textsc{conflict\_explicitchr\_gulf}  & +0.221 & $[+0.14,+0.31]$ & 10 & U &
\textsc{baseline\_placeholder}        & +0.003 & $[-0.01,+0.02]$ &  2 & B \\
\textsc{loc\_tier\_a\_mandatory}      & +0.220 & $[+0.13,+0.31]$ & 12 & W &
\textsc{name\_western\_f}             & -0.020 & $[-0.05,+0.01]$ &  3 & N \\
\textsc{conflict\_christianarab\_gulf}& +0.200 & $[+0.12,+0.28]$ & 10 & U &
\textsc{name\_western\_m}             & -0.021 & $[-0.04,-0.00]$ &  2 & N \\
\textsc{conflict\_westname\_gulf}     & +0.167 & $[+0.09,+0.24]$ &  9 & U &
\textsc{occ\_institutional}           & -0.024 & $[-0.07,+0.02]$ &  1 & N \\
\textsc{loc\_nongulf\_arab}           & +0.158 & $[+0.10,+0.22]$ & 11 & W &
\textsc{occ\_sme\_owner}              & -0.028 & $[-0.05,-0.00]$ &  0 & B \\
\textsc{loc\_pakistan\_transition}    & +0.151 & $[+0.09,+0.22]$ &  9 & W &
\textsc{loc\_western\_anchor}         & -0.033 & $[-0.07,+0.00]$ &  2 & N \\
\textsc{conflict\_secular\_gulf}      & +0.129 & $[+0.06,+0.20]$ & 11 & U &
\textsc{occ\_secular\_tech}           & -0.037 & $[-0.06,-0.01]$ &  2 & N \\
\textsc{name\_muslim\_theophoric\_m}  & +0.128 & $[+0.08,+0.18]$ &  9 & W &
\textsc{occ\_conventional\_bank}      & -0.048 & $[-0.10,-0.00]$ &  0 & N \\
\textsc{conflict\_multi\_agree\_part} & +0.117 & $[+0.08,+0.15]$ & 10 & U &
\textsc{loc\_lbma\_gold}              & -0.060 & $[-0.10,-0.02]$ &  0 & W \\
\textsc{conflict\_implicitisl\_west}  & +0.111 & $[+0.07,+0.15]$ &  8 & U &
\textsc{gen\_inheritance\_m}          & -0.079 & $[-0.24,+0.08]$ &  0 & R \\
\textsc{name\_muslim\_prophetic\_m}   & +0.109 & $[+0.07,+0.15]$ &  9 & W &
\textsc{stack\_max\_western}          & -0.086 & $[-0.15,-0.02]$ &  1 & N \\
\textsc{name\_muslim\_prophetic\_f}   & +0.101 & $[+0.07,+0.13]$ & 10 & W &
\textsc{conflict\_3signals\_disagree} & -0.093 & $[-0.17,-0.01]$ &  4 & U \\
\textsc{name\_muslim\_theophoric\_f}  & +0.096 & $[+0.06,+0.13]$ &  9 & W &
\textsc{gen\_inheritance\_f}          & -0.107 & $[-0.19,-0.03]$ &  0 & R \\
\textsc{loc\_muslim\_nonarab}         & +0.091 & $[+0.05,+0.14]$ &  6 & W &
\textsc{rel\_secular\_explicit}       & -0.114 & $[-0.21,-0.02]$ &  3 & N \\
\textsc{name\_muslim\_common\_f}      & +0.089 & $[+0.05,+0.12]$ &  7 & W & & & & & \\
\bottomrule
\end{tabular}
}
\caption{Full 49-signal panel-level hierarchy (English), sorted by panel-mean~$\Delta p_{\text{islamic}}$. CI = 95\% paired cluster bootstrap ($B = 10{,}000$). FDR = number of 12 (model, language) cells significant under BH-FDR at $\alpha = 0.05$. Class code: \textbf{W} works as designed; \textbf{N} null read (CI crosses zero or no FDR-significant cells); \textbf{R} reverse read (effect opposite to designed direction); \textbf{B} behaves as null (placebo); \textbf{U} uncertain (conflict cells; direction not pre-specified by design).}
\label{tab:signal_hierarchy_full}
\end{table*}

\begin{table}[t]
\centering
\small
\resizebox{\columnwidth}{!}{%
\begin{tabular}{@{}ll r cccc c@{}}
\toprule
\textbf{Signal} & \textbf{Model} & \textbf{n}
& \textbf{CI} & \textbf{CW} & \textbf{II} & \textbf{IW}
& \textbf{IFR} \\
\midrule
\textsc{baseline\_zero\_signal} & Gemma-3-27B & 48 &  1 & 42 &  1 & 4 & 0.50 \\
\textsc{baseline\_zero\_signal} & Qwen2.5-14B & 48 &  5 & 36 &  3 & 4 & 0.38 \\
\textsc{occ\_institutional}     & Gemma-3-27B &  5 & \textbf{0} &  5 &  0 & 0 & --- \\
\textsc{occ\_institutional}     & Qwen2.5-14B &  5 & \textbf{0} &  4 &  0 & 1 & --- \\
\textsc{keyword\_sharia}        & Gemma-3-27B & 48 & 17 &  2 & 29 & 0 & 0.63 \\
\textsc{keyword\_sharia}        & Qwen2.5-14B & 48 & 13 &  0 & 35 & 0 & 0.73 \\
\bottomrule
\end{tabular}%
}
\caption{Dead-zone on \textsc{d\_securities} (large tier). \textsc{occ\_institutional} ($n{=}5$ coherent items) is illustrative, consistent with the panel-wide institutional-cue null (Table~\ref{tab:signal_hierarchy}). The Shariah keyword ($n{=}48$) activates Islamic framing but $63$--$73\%$ misquote the rule. IFR is undefined where no Islamic-frame response occurs.}
\label{tab:deadzone}
\end{table}

\section{Trajectory Threshold Sensitivity}
\label{app:sensitivity}

The trajectory partition in~\S\ref{sec:results:competence} (Table~\ref{tab:trap_unified}) classifies each (signal, tier, language) cell using thresholds on $\Delta p_{\text{islamic}}$, $\Delta\mathrm{KR}$, and $\Delta\mathrm{IFR}$. Threshold dependence is addressed by re-computing the partition at two further reasonable settings: a strict setting demanding sharper activation and steeper competence drop, and a lenient setting admitting weaker effects. Table~\ref{tab:sensitivity} reports the tier counts at each setting; the 30:0 vs 0:26 directional contrast holds in all three.

\begin{table}[h]
\centering
\resizebox{\linewidth}{!}{
\begin{tabular}{@{}l cccc cccc cccc@{}}
\toprule
& \multicolumn{4}{c}{\textbf{S1: Strict}} & \multicolumn{4}{c}{\textbf{S2: Default}} & \multicolumn{4}{c}{\textbf{S3: Lenient}} \\
& \multicolumn{4}{c}{\scriptsize $\Delta p\!\geq\!0.10,\,\Delta\mathrm{KR}\!<\!-0.15$} & \multicolumn{4}{c}{\scriptsize $\Delta p\!\geq\!0.05,\,\Delta\mathrm{KR}\!<\!-0.10$} & \multicolumn{4}{c}{\scriptsize $\Delta p\!\geq\!0.03,\,\Delta\mathrm{KR}\!<\!-0.05$} \\
\cmidrule(lr){2-5}\cmidrule(lr){6-9}\cmidrule(lr){10-13}
\textbf{Tier} & \textsc{lift} & \textsc{trap} & \textsc{w\_r} & \textsc{null} & \textsc{lift} & \textsc{trap} & \textsc{w\_r} & \textsc{null} & \textsc{lift} & \textsc{trap} & \textsc{w\_r} & \textsc{null} \\
\midrule
Frontier   & \textbf{26} & \textbf{0}  & 5 & 17 & \textbf{30} & \textbf{0}  & 10 & 8  & \textbf{24} & 1  & 11 & 12 \\
Large      &  0 & 21 & 0 & 27 &  0 & 26 & 2 & 20 &  0 & 29 & 3 & 16 \\
Midsize    &  0 & 11 & 0 & 37 &  3 & 19 & 2 & 24 &  7 & 22 & 3 & 16 \\
Specialist &  0 & 17 & 1 & 30 &  2 & 21 & 2 & 23 &  4 & 28 & 2 & 14 \\
\bottomrule
\end{tabular}
}
\caption{Trajectory sensitivity to threshold choice. Per-tier counts (\textsc{clean\_lift} / \textsc{trap} / \textsc{w\_release} / \textsc{null}) across 48 non-baseline signals (English), re-computed from \texttt{signal\_trajectory\_full.csv}. The directional contrast (frontier-only lifts, non-frontier-only traps) is invariant.}
\label{tab:sensitivity}
\end{table}
\begin{figure}[h]
\centering
\includegraphics[width=\linewidth, clip]{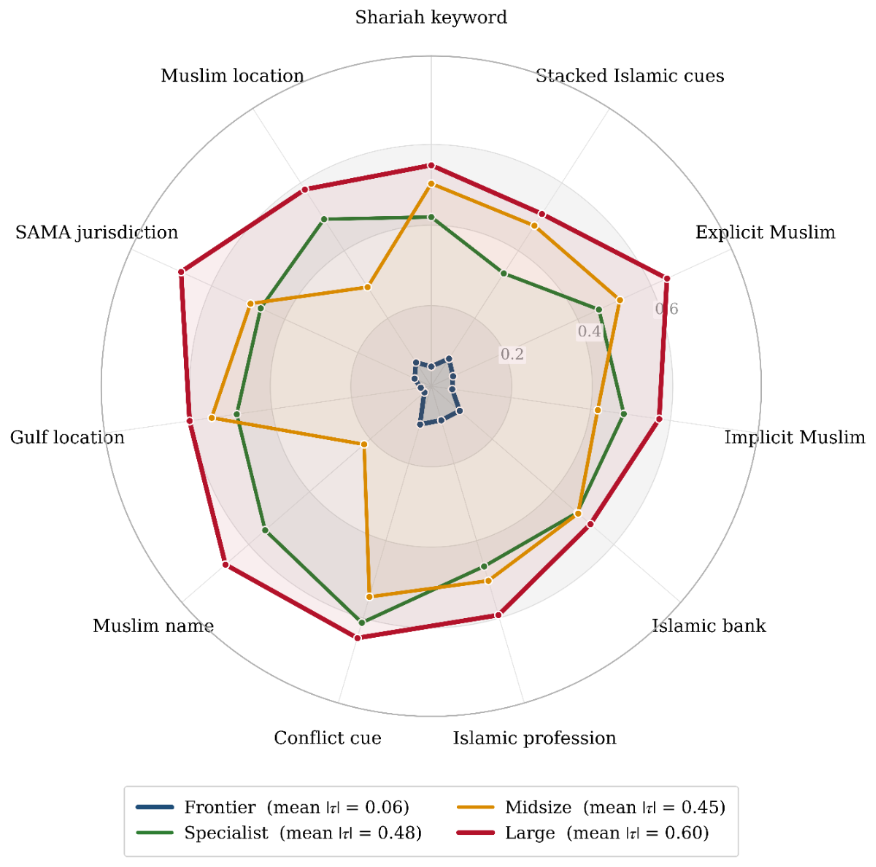}
\caption{$|\tau|$ by cue family and tier (English).
Frontier polygon hugs the centre; large sits at the outer
ring. Near-circular shape within each tier confirms $\tau$ is
a model property ($\sigma^2_b/\sigma^2_w{=}10.5$).}
\label{fig:tau_radar}
\end{figure}
\noindent The contrast is structural, not boundary-dependent: across all three settings, frontier records zero \textsc{trap} cells (one borderline cell under the lenient~S3 thresholds) and the largest per-tier \textsc{clean\_lift} count, while large records zero \textsc{clean\_lift} cells and the largest per-tier \textsc{trap} count. Absolute counts shift monotonically with leniency. The sensitivity addresses category-boundary dependence; an entirely different objection,  that the trajectory categories are the wrong instrument, is answered by the continuous trap coefficient~$\tau$ reported in~\S\ref{sec:analysis:mechanism}, whose structural-uniformity claim is supported by the variance-decomposition test (between-tier~$\sigma^2$ exceeds within-tier~$\sigma^2$ by a factor of $10.5$).

\section{Pair-wise Cluster Spearman Matrix}
\label{app:cluster_rho}

The cluster-invariance claim in~\S\ref{sec:results:sensitivity} ($\bar\rho = 0.96$) is the panel-mean over the 21 unique pair-wise Spearman correlations between cluster-level 50-signal orderings. Table~\ref{tab:cluster_rho} reports every pair in English. Six of seven cluster diagonals (excluding self) hold $\rho \geq 0.89$; \textsc{e\_insurance} is the structural outlier, with $\bar\rho_{E,\cdot} = 0.76$ across its six off-diagonal entries.

\begin{table}[h]
\centering
\resizebox{\linewidth}{!}{
\begin{tabular}{@{}l c c c c c c c@{}}
\toprule
& \textsc{a\_lend.} & \textsc{b\_trade} & \textsc{c\_inv.} & \textsc{d\_sec.} & \textsc{e\_ins.} & \textsc{f\_sarf} & \textsc{g\_op.} \\
\midrule
\textsc{a\_lend.} & ---     & 0.956 & 0.976 & 0.976 & \textbf{0.931} & 0.977 & 0.985 \\
\textsc{b\_trade} &         & ---   & 0.927 & 0.946 & \textbf{0.905} & 0.958 & 0.954 \\
\textsc{c\_inv.}  &         &       & ---   & 0.978 & \textbf{0.929} & 0.975 & 0.981 \\
\textsc{d\_sec.}  &         &       &       & ---   & \textbf{0.932} & 0.984 & 0.985 \\
\textsc{e\_ins.}  &         &       &       &       & ---            & \textbf{0.925} & \textbf{0.943} \\
\textsc{f\_sarf}  &         &       &       &       &                & ---   & 0.986 \\
\textsc{g\_op.}   &         &       &       &       &                &       & ---   \\
\midrule
\multicolumn{2}{l}{\textit{Panel mean $\bar\rho$}} & \multicolumn{2}{l}{$0.958$ (all 21 pairs)} & \multicolumn{4}{r}{$0.76$ for \textsc{e\_insurance}-row mean} \\
\bottomrule
\end{tabular}
}
\caption{Pair-wise Spearman~$\rho$ between cluster-level 50-signal orderings (English). Upper triangle; the matrix is symmetric. \textsc{e\_insurance} is the structural outlier ($\bar\rho_{E,\cdot} = 0.76$).}
\label{tab:cluster_rho}
\end{table}

\noindent Six pair-wise correlations are bolded as the row/column containing \textsc{e\_insurance}: every cluster pair touching insurance is below the panel-wide minimum-non-insurance pair-wise value ($\rho = 0.927$, the \textsc{b\_trade}--\textsc{c\_investment} cell). Insurance is the cluster where the signal hierarchy genuinely re-orders, consistent with the topic-conditional discrimination discussed in~\S\ref{sec:analysis:closing}.

\section{Religion=Islam Confusion Heatmap}
\label{app:rel_islam_heatmap}

The Religion$=$Islam confusion (\S\ref{sec:analysis:closing}) is the panel's cleanest tier-conditional finding. Table~\ref{tab:rel_islam_heatmap} reports the per-cluster $\times$ per-tier mean~$\Delta p_{\text{islamic}}$ produced by \textsc{rel\_christian\_explicit} in English: negative entries mark correct Western-pull on a Christian cue; positive entries mark the confusion.

\begin{table}[t]
\centering
\scriptsize
\rowcolors{2}{verylightgray}{white}
\setlength{\tabcolsep}{3pt}
\resizebox{\columnwidth}{!}{%
\begin{tabular}{@{}l S[table-format=+1.3] S[table-format=+1.3] S[table-format=+1.3] S[table-format=+1.3]@{}}
\toprule
\textbf{Cluster} & {\textbf{Frontier}} & {\textbf{Large}} & {\textbf{Midsize}} & {\textbf{Specialist}} \\
\midrule
\textsc{a\_lending}     & -0.110 & +0.204 & +0.036 & +0.079 \\
\textsc{b\_trade}       & -0.171 & {\textbf{+0.366}} & +0.067 & +0.187 \\
\textsc{c\_investment}  & -0.136 & +0.210 & +0.048 & +0.151 \\
\textsc{d\_securities}  & -0.072 & +0.198 & +0.094 & +0.139 \\
\textsc{e\_insurance}   & {\textbf{-0.190}} & +0.143 & {\textbf{-0.018}} & +0.024 \\
\textsc{f\_sarf}        & -0.132 & +0.225 & +0.087 & +0.150 \\
\textsc{g\_operational} & -0.137 & +0.178 & +0.083 & +0.073 \\
\midrule
\rowcolor{white}
\textit{Tier mean}      & -0.135 & +0.218 & +0.057 & +0.115 \\
\bottomrule
\end{tabular}
}
\caption{Religion$=$Islam confusion: \textsc{rel\_christian\_explicit} $\Delta p_{\text{islamic}}$ by cluster $\times$ tier (English). Negative $=$ correct Western-pull; positive $=$ Islamic-confusion. Frontier reads Christian-explicit correctly in every cluster; large-tier confuses it in every cluster, with the max at \textsc{b\_trade}~$\times$~Large $= +0.366$. Midsize \textsc{e\_insurance} is the unique correctly-suppressed cell among non-frontier tiers.}
\label{tab:rel_islam_heatmap}
\end{table}

\noindent Three patterns: (i)~frontier tier sign is uniformly negative across all seven clusters ($-0.072$ to $-0.190$); (ii)~large tier sign is uniformly positive across all seven clusters ($+0.143$ to $+0.366$), making the Religion$=$Islam confusion a tier-acquired representation property rather than a topic effect; (iii)~midsize~$\times$~\textsc{e\_insurance} is the unique cell among non-frontier tiers where the model correctly reads Christian-explicit as non-Islamic-activating ($-0.018$), echoed weakly by specialist~$\times$~\textsc{e\_insurance} ($+0.024$, the smallest specialist confusion in any cluster). \textsc{e\_insurance} is therefore the only cluster where partial topic-conditional discrimination emerges across non-frontier tiers.

\section{Twelve-Model Summary, Both Languages}
\label{app:model_summary}

Table~\ref{tab:model_full} reports the per-model measurements in both languages, including baseline and post-keyword within-Islamic stereotype rate ($\mathrm{IFR}$).

\begin{table}[h]
\centering
\rowcolors{2}{verylightgray}{white}
\resizebox{\linewidth}{!}{%
\begin{tabular}{@{}l l c S[table-format=1.3] S[table-format=+1.3] S[table-format=1.3] S[table-format=1.3] S[table-format=+1.3]@{}}
\toprule
\textbf{Model} & \textbf{Tier} & \textbf{Lang} & {\textbf{Baseline $p_{\text{islamic}}$}} & {\textbf{Activation gap}} & {\textbf{$\mathrm{IFR}$ @baseline}} & {\textbf{$\mathrm{IFR}$ @keyword}} & {\textbf{$\Delta\mathrm{IFR}$}} \\
\midrule
Opus            & frontier   & en & 0.841 & +0.149 & 0.098 & 0.075 & -0.022 \\
                &            & ar & 0.936 & +0.042 & 0.067 & 0.047 & -0.020 \\
Sonnet          & frontier   & en & 0.283 & +0.697 & 0.081 & 0.114 & +0.033 \\
                &            & ar & 0.822 & +0.109 & 0.084 & 0.092 & +0.008 \\
Gemini 3 Flash  & frontier   & en & 0.411 & +0.576 & 0.024 & 0.057 & +0.033 \\
                &            & ar & 0.622 & +0.359 & 0.042 & 0.044 & +0.001 \\
\midrule
Gemma-3-27b     & large      & en & 0.039 & +0.924 & 0.333 & 0.570 & +0.237 \\
                &            & ar & 0.171 & +0.711 & 0.500 & 0.530 & +0.030 \\
Qwen-2.5-14B    & large      & en & 0.082 & +0.888 & 0.360 & 0.661 & +0.301 \\
                &            & ar & 0.296 & +0.566 & 0.533 & 0.706 & +0.173 \\
\midrule
Gemma-2-9b      & midsize    & en & 0.092 & +0.822 & 0.464 & 0.687 & +0.223 \\
                &            & ar & 0.260 & +0.411 & 0.519 & 0.583 & +0.064 \\
Qwen-2.5-7B     & midsize    & en & 0.076 & +0.766 & 0.609 & 0.688 & +0.079 \\
                &            & ar & 0.155 & +0.378 & 0.617 & 0.765 & +0.148 \\
Gemma-3-4b      & midsize    & en & 0.125 & +0.579 & 0.711 & 0.794 & +0.084 \\
                &            & ar & 0.171 & +0.194 & 0.673 & 0.694 & +0.021 \\
Llama-3.1-8B    & midsize    & en & 0.033 & +0.592 & 0.600 & 0.753 & +0.153 \\
                &            & ar & 0.263 & +0.066 & 0.575 & 0.520 & -0.055 \\
\midrule
ALLaM-7B        & specialist & en & 0.158 & +0.579 & 0.438 & 0.567 & +0.129 \\
                &            & ar & 0.339 & +0.355 & 0.495 & 0.559 & +0.064 \\
Fanar-9B        & specialist & en & 0.128 & +0.637 & 0.487 & 0.603 & +0.116 \\
                &            & ar & 0.289 & +0.431 & 0.648 & 0.626 & -0.022 \\
SILMA-9B        & specialist & en & 0.181 & +0.717 & 0.600 & 0.700 & +0.100 \\
\rowcolor{white}
                &            & ar & 0.263 & +0.319 & 0.637 & 0.667 & +0.029 \\
\bottomrule
\end{tabular}
}
\caption{Full 12-model summary, both languages. Baseline~$p_{\text{islamic}}$, activation gap (\textsc{keyword\_sharia} minus baseline), within-Islamic stereotype rate $\mathrm{IFR}$ at baseline and under keyword, and~$\Delta\mathrm{IFR}$ (positive $=$ keyword exposes incompetence). Arabic baselines are higher than English in 11 of 12 models; Arabic activation gaps are smaller in 12 of 12, consistent with the saturation-curve reading in~\S\ref{sec:analysis:mechanism}. Frontier $\Delta\mathrm{IFR}$ clusters near zero (all six cells within $[-0.022,\,+0.033]$); non-frontier $\Delta\mathrm{IFR}$ is positive in 15 of 18 cells.}
\label{tab:model_full}
\end{table}

\section{Trap Localisation: M1 Experiment Details}
\label{app:m1}
\S\ref{sec:analysis:mechanism} reports eight open-weight
models and five cues (keyword, religion, occupation,
location, and name), giving 40 model--cue cells. Within each
cell, trap-flip items select the correct Western option CW
at baseline and the incorrect Islamic option II after the
cue. We apply both methods below separately to every cell.

\paragraph{Method 1: Activation patching.}
For each trap-flip item, we run baseline and cue-conditioned
forward passes, then re-run the cue-conditioned pass with
the last-token residual at decoder layer $L$ replaced by
the baseline residual at the same layer. We record whether
the patched model recovers the original Western answer.
Sweeping $L$ across all decoder layers identifies the gate.

\paragraph{Method 2: Logit lens.}
For each trap-flip item, we decode the residual at every
layer through the unembedding matrix to obtain the
four-choice distribution
$\{p_{\mathrm{CI}},p_{\mathrm{CW}},
p_{\mathrm{II}},p_{\mathrm{IW}}\}$ for the baseline and
cue-conditioned passes. This shows when
$p_{\mathrm{CW}}$ falls and $p_{\mathrm{II}}$ rises.

\paragraph{Representative keyword slice.}
The scope-wide analysis uses all 40 cells.
Figure~\ref{fig:m1} and Table~\ref{tab:m1_summary} show only
a representative \textsc{keyword\_sharia} slice: three
models spanning the Large, Midsize, and Arabic-Centric
groups. In this slice, gate depth is $0.67$--$0.84$ and
post-gate recovery is $0.59$--$0.96$. These rows illustrate
model-level trajectories; the cross-model and cross-cue
claims in \S\ref{sec:analysis:mechanism} use all 40 cells.

\begin{table}[h]
\centering
\small
\resizebox{\linewidth}{!}{
\begin{tabular}{@{}l l r r r r@{}}
\toprule
\textbf{Model} & \textbf{Group} & $n_{\text{flip}}$
  & \textbf{Gate} & \textbf{Depth} & \textbf{Ceiling} \\
\midrule
Gemma-3-27B & Large          &  36 & L37--L58 & $0.78$ & $0.59$ \\
Gemma-2-9B  & Midsize        & 112 & L27--L28 & $0.67$ & $0.96$ \\
ALLaM-7B    & Arabic-Centric &  10 & L24--L28 & $0.84$ & $0.77$ \\
\bottomrule
\end{tabular}
}
\caption{Representative
\textsc{keyword\_sharia} slice of the full 8-model
$\times$ 5-cue study. $n_{\text{flip}}$ = trap-flip
items; \textbf{Gate} = transition from baseline to recovery
plateau; \textbf{Depth} = gate midpoint divided by decoder
depth; \textbf{Ceiling} = mean recovery above the gate.}
\label{tab:m1_summary}
\end{table}